\documentclass{article}

\PassOptionsToPackage{table}{xcolor}

\usepackage{microtype}
\usepackage{graphicx}
\usepackage{subfigure}
\usepackage{booktabs} % for professional tables

\usepackage{hyperref}

\usepackage[accepted]{sty/icml2026}

\usepackage{amsmath}
\usepackage{amssymb}
\usepackage{mathtools}
\usepackage{amsthm}

\usepackage{multirow} 

\usepackage[capitalize,noabbrev]{cleveref}

\usepackage{amssymb}    % 数学符号，如 \varnothing
\usepackage{booktabs}   % 表格横线命令，如 \toprule, \midrule, \bottomrule
\usepackage{multirow}   % 跨行单元格命令，如 \multirow
\usepackage{makecell}   % 表格单元格内换行或复杂格式，如 \makecell
\usepackage{pifont}     % 供了很多特殊符号（比如打勾、打叉）
\usepackage{xcolor}     % 提供 \color (颜色支持)
\usepackage[table]{xcolor}
\definecolor{myblue}{HTML}{DAEFF9}
\definecolor{revisioncolor}{HTML}{000000}
\newcommand{\revise}[1]{\textcolor{revisioncolor}{#1}}

\usepackage{colortbl}
\definecolor{mycolumn}{HTML}{F5F5F5}
\usepackage{adjustbox}
\newcommand{\cmark}{\ding{51}}
\newcommand{\xmark}{\textcolor{gray}{\ding{55}}}
\definecolor{rowgray}{RGB}{245,245,245}
\definecolor{rowblue}{RGB}{219,239,252}
\usepackage{bm} % 加粗符号
\usepackage{bbm} % 提供 \mathbbm
\usepackage{threeparttable}
\usepackage{pdfpages}
\usepackage{etoc}
\theoremstyle{plain}

\theoremstyle{definition}

\theoremstyle{remark}

\usepackage[textsize=tiny]{todonotes}

\icmltitlerunning{Linguistic Relative Policy Optimization for Video Anomaly Reasoning}

\begin{document}

\twocolumn[
\icmltitle{Linguistic Relative Policy Optimization for Video Anomaly Reasoning}



\icmlsetsymbol{equal}{*}

\begin{icmlauthorlist}
\icmlauthor{Jiaxu Leng}{cqupt,cqai}
\icmlauthor{Jiankang Zheng}{cqupt}
\icmlauthor{Mengjingcheng Mo}{cqupt}
\icmlauthor{Zhanjie Wu}{cqupt}
\icmlauthor{Haosheng Chen}{cqupt}
\icmlauthor{Ji Gan}{cqupt}
\icmlauthor{Xinbo Gao}{cqupt}
\end{icmlauthorlist}

\icmlaffiliation{cqupt}{School of Computer Science and Technology, Chongqing University of Posts and Telecommunications, Chongqing, China}
\icmlaffiliation{cqai}{Chongqing College of Artificial Intelligence, Chongqing, China}

\icmlcorrespondingauthor{Xinbo Gao}{gaoxb@cqupt.edu.cn}


\icmlkeywords{Machine Learning, ICML}

\vskip 0.3in
]

% 脚注
\printAffiliationsAndNotice{} 

\begin{abstract}
\label{sec:abstract}

Video anomaly detection (VAD) with multimodal large language models has shown strong potential, yet most existing methods still depend on large-scale annotations or expert-designed priors, limiting their ability to acquire anomaly knowledge with as little human intervention as possible. To address this, we propose Linguistic Relative Policy Optimization (LRPO), which distills group-relative semantic advantages from multiple reasoning trajectories into a linguistically expressed anomaly experience prior, and adapts the model by injecting this prior into the context to steer its output distribution without any parameter updates. LRPO builds two complementary experience representations: general experience captures transferable anomaly preferences across scenarios, while scenario experience models context-dependent anomaly rules for targeted refinement. To further improve the learned experience, we introduce an anomaly alignment reward that guides trajectory optimization to match human risk preferences and reinforce temporally grounded reasoning. Extensive experiments on XD-Violence, UCF-Crime, and UBnormal demonstrate that LRPO significantly outperforms existing state-of-the-art methods under tuning-free settings.

\end{abstract}

% Main Sections
% Introduction
\section{Introduction}
\label{sec:introduction}
\begin{figure}[!htb]
\centerline{\hspace{-0.04\linewidth}\includegraphics[width=1.0\linewidth, keepaspectratio]{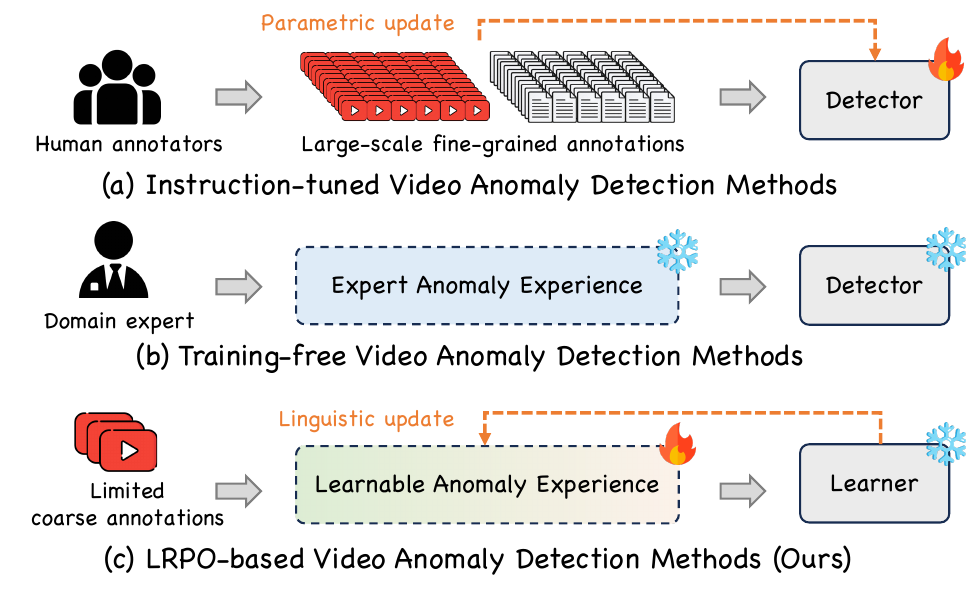}}
% \vspace{-8pt}
\caption{Recent research in VAD can be categorized into two types: (a) instruction-tuned VAD, which aligns anomaly knowledge by updating model parameters with large-scale fine-grained annotations; and (b) training-free VAD, which freezes model parameters and relies on expert priors to guide anomaly reasoning. We propose LRPO-based VAD: (c) it also freezes model parameters but learns linguistic anomaly experience from limited coarse annotations, substantially reducing human involvement.}
\label{fig:motivation}
\vspace{-6pt}
\end{figure}

Video Anomaly Detection (VAD) automatically detects and localizes rare, unexpected, or hazardous events in video streams, a capability that is critical for safety-sensitive video analysis~\cite{sultani2018real,mo2024nexusad}. Despite strong performance of conventional semi-supervised and weakly supervised VAD methods~\cite{cao2024context,zhang2024multi,wu2024vadclip,huang2025multimodal, meng2025audio} under the training distribution, real-world deployments are often highly diverse and commonly exhibit substantial domain shifts between training and test environments~\cite{li2023logical,li2025video,li2025shape,mo2026a2seek}. Consequently, their performance typically degrades severely under out-of-distribution environments. Recently, multimodal large language models (MLLMs) have demonstrated strong generalization across a broad range of challenging visual tasks, suggesting a promising direction for enhancing the generalization of VAD under distribution shifts.

As illustrated in Figure~\ref{fig:motivation}, existing MLLM-based VAD methods broadly fall into two paradigms:
(1) Instruction-tuned VAD. These methods~\cite{tang2024hawk,zhang2025holmes} fine-tune MLLMs to align anomaly semantics with the VAD task; however, they typically require large-scale, fine-grained annotations and thus incur non-trivial human annotation costs.
(2) Training-free VAD. These methods~\cite{zanella2024harnessing,li2026vadtree} freeze pretrained parameters to preserve the foundation model’s generalization; however, they typically depend on hand-crafted reasoning pipelines based on expert knowledge rather than data-driven feedback, hindering automated iteration and necessitating manual tuning.
\textbf{Overall, both paradigms require substantial human intervention, either large-scale annotation or manual pipeline design and tuning, hindering generalizable anomaly knowledge acquisition under distribution shifts.}
\begin{table}[t]
\centering
% \vspace{-4px}
\caption{Performance impact of different experience types on XD-Violence using InternVL3\_5-8B as the base model.}
\label{tab:abla_exp}
\setlength{\tabcolsep}{10pt}
% \small
\begin{tabular}{l c}
\toprule
\textbf{Experience Type} & \textbf{AP (\%)} \\
\midrule
without experience & 59.93\\
with manually designed experience & 68.47\\
with single-trajectory learned experience & 66.78 \\
with \textbf{LRPO-learned experience (Ours)} & \textbf{73.17}\\
\bottomrule
\end{tabular}

% \vspace{-10pt}
\end{table}

% % Preamble
% % \usepackage{booktabs}
% % \usepackage[table]{xcolor}
% \definecolor{rowblue}{RGB}{219,239,252}

% \begin{table}[t]
% \centering
% \caption{Performance impact of different experience types on the XD-Violence dataset.}
% \vspace{2px}
% \label{tab:abla_exp}
% \setlength{\tabcolsep}{10pt}
% \resizebox{0.85\linewidth}{!}{
% \begin{tabular}{l c}
% \toprule
% \textbf{Experience Type} & \textbf{AP (\%)} \\
% \midrule
% \multicolumn{2}{l}{\textit{Base model: InternVL3\_5-8B}} \\
% \hspace{1em} without experience & 59.93\\
% \hspace{1em} with manually designed experience & 68.47\\
% \hspace{1em} with pointwise-learned experience & 68.48 \\
% \hspace{1em} with \textbf{LRPO-learned experience (Ours)} & \textbf{73.17}\\

% % \hspace{1em} w/o experience & 59.93\\
% % \hspace{1em} w/  manually designed experience & 68.47\\
% % \hspace{1em} w/  pointwise-learned experience & 68.48 \\
% % \hspace{1em} w/  \textbf{LRPO-learned experience (Ours)} & \textbf{73.17}\\
% \bottomrule
% \end{tabular}
% }
% \vspace{-10pt}
% \end{table}

By contrast, humans can often learn anomaly discrimination from limited data by leveraging broad commonsense knowledge, requiring only calibration of what constitutes an anomaly under task-specific risk preferences. Consistent with this, Table~\ref{tab:abla_exp} compares several forms of anomaly-related experience, \revise{including human-written anomaly judgment rules and experience learned from correctness feedback on a single reasoning trajectory.} The results show that directly applying a frozen vision–language model (VLM) to VAD results in poor performance, whereas providing anomaly-related experience in the context improves performance markedly. These results indicate that pretrained VLMs already encode rich general knowledge, and that the main bottleneck is the lack of explicit anomaly experience rather than limited model capacity.

Based on these observations, we propose \textbf{Linguistic Relative Policy Optimization (LRPO)}, which learns linguistically expressed anomaly experience from limited data and injects it into the model input context to improve anomaly reasoning in a tuning-free manner. A key challenge is that anomaly perception is inherently subjective and context-dependent: the same video may admit different judgments under different risk preferences, making optimization toward a single absolute label noisy and unstable. To address this, LRPO draws on relative optimization from reinforcement learning to iteratively learn and refine anomaly experience. Concretely, a \textit{Learner} VLM generates multiple reasoning trajectories per sample to cover diverse anomaly preferences. These trajectories are then assigned reward scores, and an \textit{Optimizer} LLM performs group-wise reflection to extract semantic advantages. The extracted advantages are distilled into anomaly experience and injected as a dynamically updated contextual signal, progressively steering the Learner’s outputs toward the target risk preferences. \revise{Different from conventional reinforcement learning that optimizes model parameters, LRPO instantiates relative optimization in a language-editable experience space, where a persistent anomaly experience repository is updated by distilling semantic differences between high- and low-reward reasoning trajectories.}

Within this framework, we build two complementary experience representations: \textbf{general experience}, learned by LRPO to capture transferable anomaly preferences across scenarios, and \textbf{scenario experience}, constructed by the VLM with weak-label prompting to encode context-dependent anomaly rules and expand as more data become available. Together, they support effective scenario adaptation while preserving generalization.
To further improve the learned anomaly experience, we introduce an \textbf{anomaly alignment reward} with two complementary terms. The anomaly preference reward first aligns the learner’s judgments with human risk preferences by using the scenario experience of the current sample as a positive reference and contrasting it with LLM-generated perturbed outputs. Building on this, the anomaly temporal dependency reward further promotes temporally grounded reasoning by contrasting performance on temporally ordered frames with randomly permuted inputs, encouraging step-by-step inference based on temporal evidence.

Our main contributions are summarized as follows:
1) We propose LRPO, a tuning-free framework for VAD that distills group-relative semantic advantages from multiple reasoning trajectories into a linguistically expressed anomaly experience prior, and adapts the model by injecting this prior into the context to steer its output distribution. 
2) We build general and scenario experience to capture transferable anomaly preferences and context-dependent anomaly rules.
3) We introduce an anomaly alignment reward to optimize experience, aligning it with human risk preferences and reinforcing temporally grounded reasoning.
4) We achieve state-of-the-art tuning-free performance on XD-Violence, UCF-Crime, and UBnormal, with strong generalization across models and datasets.

% \textbf{Conflict of Interest Disclosure.}
% The authors declare no financial conflicts of interest related to this work.

% Related Work
\section{Related Work}

\textbf{Tuning-based VAD.}
These methods update model parameters with supervision signals of varying granularity to align anomaly-related semantics with decision boundaries, including semi-supervised VAD~\cite{yan2023feature, cao2024context, zhang2024video, zhang2024multi, zhu2024advancing}, weakly-supervised VAD~\cite{sultani2018real, tian2021weakly, li2022scale, lv2023unbiased, shi2023abnormal, chen2024prompt, wu2024vadclip, huang2025multimodal, meng2025audio, leng2025piercingeye}, and instruction-tuned VAD~\cite{du2024uncovering, tang2024hawk, zhang2024holmes, zhang2025holmes}. 
For example, \cite{rai2024video} propose spatio-temporal pseudo-anomaly generation to synthesize hard cases and enhance anomaly representations and detection robustness; \cite{huangex} propose Ex-VAD, which performs explainable fine-grained anomaly detection based on vision-language models; \cite{zhang2025holmes} combine an anomaly-oriented temporal sampler with an instruction-tuned MLLM for anomaly recognition; \revise{\cite{huang2026vad} introduce Vad-R1, a representative framework that strengthens MLLM-based VAD with explicit anomaly reasoning and perception-to-cognition supervision.} 
Although these methods typically perform strongly within the training domain, semi- and weakly-supervised methods can degrade substantially under out-of-distribution scenarios or unseen anomaly types; meanwhile, instruction-tuned methods often require large-scale fine-grained data and manually constructed instructions, incurring high human annotation costs.

\textbf{Tuning-free VAD.}
These methods freeze the parameters of foundation models and directly leverage the pretrained capabilities of MLLMs for anomaly inference~\cite{zanella2024harnessing, yang2024follow, shao2025eventvad, cai2026headhunt, cai2025hiprobe, li2026vadtree, lin2026unified, yang2026panda}. 
For example, \cite{zanella2024harnessing} improve anomaly scoring via cross-modal alignment while suppressing noisy descriptions; \cite{shao2025eventvad} segment videos into semantically coherent events and adopt hierarchical prompting for LLMs to enable anomaly localization; \cite{yang2026panda} propose an agent-based generalist VAD method that adapts to new scenes and unseen anomaly types through a closed-loop reasoning process. 
As they avoid domain-specific tuning, these methods largely preserve the generalization of foundation models and can be deployed quickly.
However, they typically depend on hand-crafted reasoning pipelines or predefined agentic procedures grounded in expert knowledge rather than data-driven feedback, which hinders automated iteration and often necessitates manual tuning.
Motivated by recent verbalized learning techniques~\cite{yuksekgonul2025optimizing,xiao2024verbalized,cai2025training}, we propose LRPO to address the above limitations. 
Unlike existing verbalized-learning-based VAD~\cite{ye2025vera}, which mainly learns guiding questions and uses them at inference time to steer the model toward local cues of anomaly patterns, LRPO explicitly models and iteratively optimizes preference experiences for anomaly judgment, thereby elevating anomaly decisions from question-guided cue focusing to experience-driven anomaly reasoning. 

\section{Method}
\begin{figure*}[!ht]
% \vspace{-2pt}

\centering
\includegraphics[width=\textwidth]{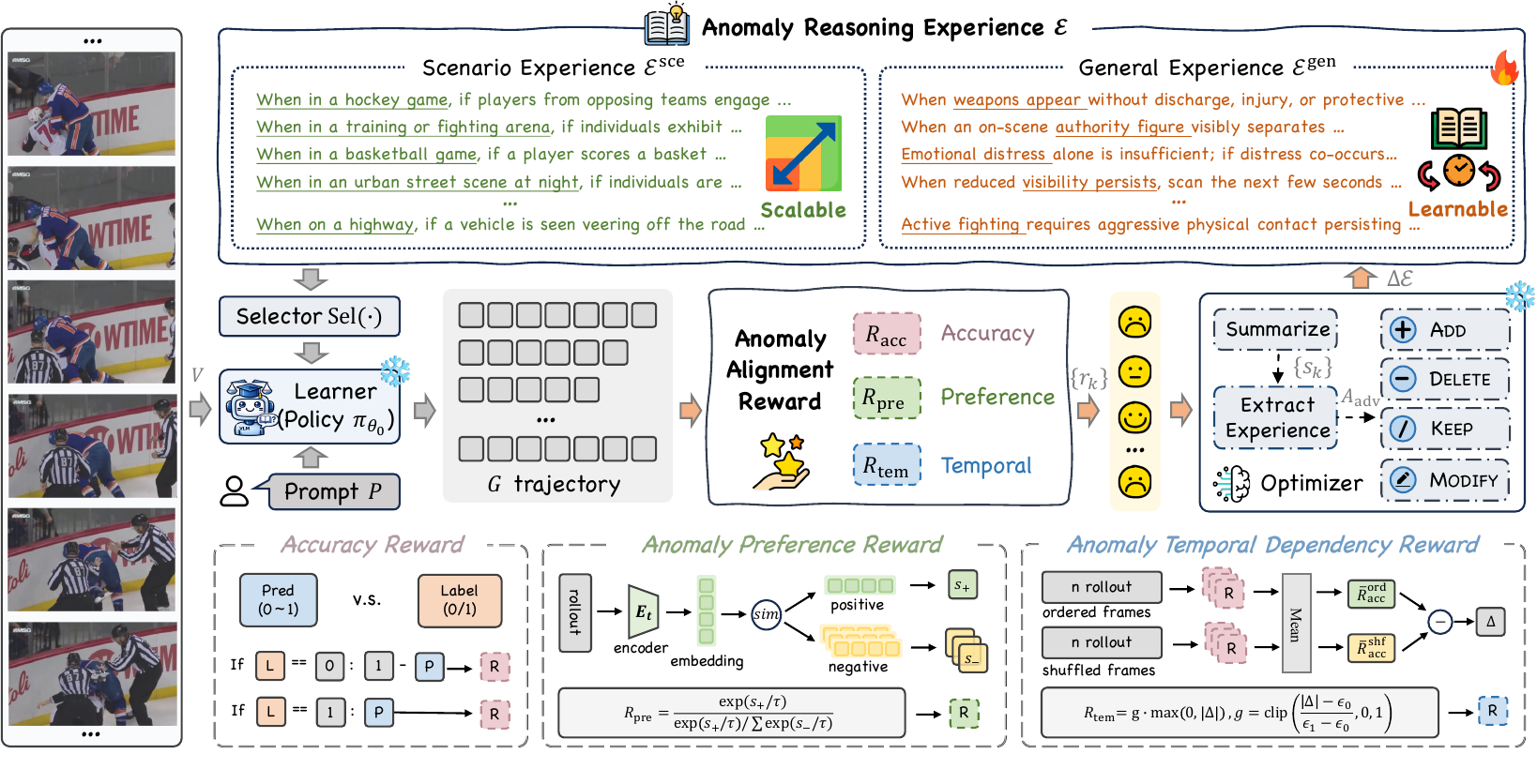} % Reduce the figure size so that it is slightly narrower than the column.
\vspace{-20pt}
\caption{
    Pipeline of LRPO. LRPO leverages linguistic interaction between a \emph{Learner} VLM and an \emph{Optimizer} LLM to iteratively optimize the general anomaly experience under reward constraints, and injects the experience as context to adjust the Learner’s output distribution, thereby adapting it to the video anomaly detection task.
    }
\label{fig:pipeline}
\end{figure*}

\subsection{Problem Formulation}
\revise{Given a video $V$,} video anomaly detection (VAD) aims to predict frame-level anomaly scores and localize anomalous temporal segments. Since anomalies are rare and frame-level annotations are costly, training data are typically provided with only video-level weak labels. We denote the training set as $\mathcal{D}=\{(V^{(j)},Y^{(j)})\}_{j=1}^{N}$, where $Y^{(j)}\in\{0,1\}$ indicates whether video $V^{(j)}$ contains anomalous events. To reduce computation while preserving key temporal evidence, we adopt the sampler in~\cite{zhang2025holmes} to \revise{convert each video into an $M$-frame key sequence $\tilde{V}^{(j)}$,} yielding $\tilde{\mathcal{D}}=\{(\tilde{V}^{(j)},Y^{(j)})\}_{j=1}^{N}$.
We optimize a vision-language model (VLM) on $\tilde{\mathcal{D}}$ for VAD, where the optimization target is not the model parameters but an editable linguistic anomaly experience \revise{repository.}
Concretely, given the sampled key-frame sequence $\tilde{V}$ and a task prompt $P$, a frozen VLM with fixed parameters $\theta_0$ induces a conditional output distribution $o\sim \pi_{\theta_0}(\cdot\mid \tilde{V},P)$, where $o$ is a complete anomaly reasoning output. Unlike conventional paradigms that update parameters $\theta$ to directly modify $\pi_{\theta}$, we maintain an editable experience repository $\mathcal{E}$ and select an experience subset $\mathcal{E}(\tilde{V})$ as injected context, thereby modulating the conditional distribution to $o\sim \pi_{\theta_0}(\cdot\mid \tilde{V},P,\mathcal{E}(\tilde{V}))$. LRPO then iteratively updates $\mathcal{E}$ to continuously steer the output distribution of $\pi_{\theta_0}$, enabling adaptation to the VAD task and aligned risk preferences without updating any VLM parameters.

\subsection{Two Complementary Experience Representations}
\label{sec:two_exp_rep}
\revise{Anomaly reasoning requires both general reasoning principles for guidance and scene-related experience for calibrating specific anomaly decision boundaries.}
\revise{Therefore, LRPO adopts a two-level experience representation consisting of generic experience and scenario experience, which are maintained in an editable linguistic anomaly experience repository} $\mathcal{E}=\{\mathcal{E}^{\mathrm{gen}},\mathcal{E}^{\mathrm{sce}}\}$. The generic experience $\mathcal{E}^{\mathrm{gen}}=\{e_i^{\mathrm{gen}}\}$ is iteratively distilled by LRPO during training from multi-trajectory feedback, serving as a transferable prior for anomaly reasoning across scenes. In contrast, the scenario experience $\mathcal{E}^{\mathrm{sce}}=\{e_i^{\mathrm{sce}}\}$ explicitly characterizes scene-specific normal patterns and anomaly boundaries, complementing the generic prior when scene-dependent variations render the general experience insufficient. During optimization or inference, LRPO constructs an experience context $\mathcal{E}(\tilde{V})$ for an input video via a selector $\mathrm{Sel}(\cdot)$, which injects the generic experience $\mathcal{E}^{\mathrm{gen}}$ together with relevant scenario experience selected from $\mathcal{E}^{\mathrm{sce}}$, thereby achieving scene adaptivity while preserving transferability. (Details of the selector are provided in \S\ref{sec:inf_exp_sel}.)

\textbf{Scenario Experience Construction.}
Scenario experience can be constructed at scale under weak supervision. Given a training sample $(\tilde{V}^{(j)},Y^{(j)})$, we use a predefined rule template $P_{\mathrm{sce}}$ to prompt the frozen VLM to generate a corresponding entry, e.g.,
\emph{``When in \textless scene type\textgreater, if \textless cue/event\textgreater happens (or \textless object\textgreater appears), you should judge it as \textless anomaly type\textgreater.''}
\revise{Here, the weak label provides the target category, i.e., the normal class or a specific anomaly category such as fire or explosion, while the VLM only supplements the scene type and visual/event cues from the sampled video to fill the remaining slots.}
\revise{In this way, scenario experience generation is constrained by weak supervision while avoiding manual enumeration of all combinations of scenes, cues, and anomaly categories.}
Formally, we generate $e_{\mathrm{sce}}^{(j)}=\Phi_{\mathrm{VLM}}\!(\tilde{V}^{(j)},P_{\mathrm{sce}},Y^{(j)})$ and write it into the repository via $\mathcal{E}^{\mathrm{sce}}\leftarrow \mathcal{E}^{\mathrm{sce}}\cup\{e_{\mathrm{sce}}^{(j)}\}$. To support subsequent retrieval, we build offline indices for each scenario experience entry. Let $E_v(\cdot)$ and $E_t(\cdot)$ denote the visual and text encoders, respectively. For $e_{\mathrm{sce}}^{(j)}$ and its associated clip $\tilde{V}^{(j)}$, we compute $\mathbf{k}_{v}^{(j)}=E_v(\tilde{V}^{(j)})$ and $\mathbf{k}_{t}^{(j)}=E_t(e_{\mathrm{sce}}^{(j)})$, and store them in the scenario index set $\mathcal{K}^{\mathrm{sce}}=\{(e_i^{\mathrm{sce}},\mathbf{k}_{v,i},\mathbf{k}_{t,i})\}$. As data accumulate, $\mathcal{E}^{\mathrm{sce}}$ can be incrementally expanded to cover more scenes and anomaly types.

\subsection{LRPO Optimization Pipeline}
As illustrated in Figure~\ref{fig:pipeline}, LRPO iteratively updates linguistic anomaly experience through interactions between a learner VLM and an optimizer LLM, under the constraint of reward signals. This progressively modulates the learner's output distribution and adapts it to the VAD task. For a theoretical analysis of LRPO's optimization process, please refer to Appendix~\ref{sec:app_theory}.

\textbf{Trajectory Sampling.}
To explore diverse plausible anomaly explanations and risk preferences for the same input, we perform group sampling to obtain $G$ trajectories from the learner. \revise{These parallel trajectories provide comparable candidates for reward-based ranking and subsequent semantic advantage distillation.} Specifically, we denote the learner VLM as a policy $\pi_{\theta_0}$, where $\theta_0$ are fixed pretrained parameters kept frozen throughout LRPO. For the current sample $(\tilde{V}^{(j)},Y^{(j)})$, the selector constructs an experience context $\mathcal{E}^{(j)}=\mathrm{Sel}(\mathcal{E},\tilde{V}^{(j)})$ from the experience repository, and we sample a group of outputs from the conditional distribution $\pi_{\theta_0}(\cdot\mid \tilde{V}^{(j)},P,\mathcal{E}^{(j)})$:
\begin{equation}
\mathcal{O}^{(j)}=\{o^{(j)}_k\}_{k=1}^{G},\qquad
o^{(j)}_k \sim \pi_{\theta_0}(\cdot\mid \tilde{V}^{(j)},P,\mathcal{E}^{(j)}),
\end{equation}
where each $o^{(j)}_k$ is a complete anomaly reasoning output.

\textbf{Reward Computation.}
To constrain experience learning and provide comparable supervision signals, we evaluate each trajectory in the output group with an anomaly alignment reward $R(\cdot)$, yielding scalar rewards $r_k^{(j)}=R(\tilde{V}^{(j)},Y^{(j)},o_k^{(j)})$. The reward design is detailed in \S\ref{sec:ano_ali_rew}.

\textbf{Semantic Advantage Distillation.}
Although the reward induces an ordering over sampled outputs, it is a scalar signal and does not specify \emph{what to change} in the experience repository.
We therefore use the optimizer to perform reflective group-wise comparisons and distill semantic advantages.
Concretely, the optimizer first produces a structured summary for each output using a summarization template $P_{\mathrm{sum}}$:
$s_k^{(j)}=\Phi_{\mathrm{opt}}(\tilde{V}^{(j)},P_{\mathrm{sum}},o_k^{(j)})$.
Given the summaries $\{s_k^{(j)}\}_{k=1}^{G}$, rewards $\{r_k^{(j)}\}_{k=1}^{G}$, and the current experience context $\mathcal{E}^{(j)}$, it contrasts high- and low-reward outputs to extract semantic advantage items with a prompt $P_{\mathrm{adv}}$:
\begin{equation}
A_{\mathrm{adv}}^{(j)}=\Phi_{\mathrm{opt}}(\tilde{V}^{(j)}, P_{\mathrm{adv}},\{s_k^{(j)},r_k^{(j)}\}_{k=1}^{G},\mathcal{E}^{(j)}),
\end{equation}
where $A_{\mathrm{adv}}^{(j)}$ is a compact natural-language list of anomaly cues, risk boundaries, and reasoning principles associated with higher rewards, which directly guides subsequent experience updates.

\textbf{Experience Optimization.}
Given $A_{\mathrm{adv}}^{(j)}$ and the experience repository $\mathcal{E}$, the optimizer further generates an executable edit instruction set $\Delta\mathcal{E}^{(j)}$ composed of operations such as \textsc{Add}/\textsc{Modify}/\textsc{Delete}/\textsc{Keep}. \revise{Specifically, these operations edit individual experience entries rather than rewriting the whole repository. \textsc{Add} writes a newly distilled transferable rule when the semantic advantages reveal missing anomaly cues or decision boundaries; \textsc{Modify} updates a specified entry when it is partially useful but needs correction or refinement; \textsc{Delete} removes duplicated entries or entries contradicted by reward feedback; and \textsc{Keep} preserves useful entries to avoid unnecessary changes.} We then update the repository via $\mathcal{E}\leftarrow \mathrm{Update}(\mathcal{E},\Delta\mathcal{E}^{(j)})$ and proceed to the next iteration. Since experience is continuously injected as a contextual prior, the evolution of $\mathcal{E}$ directly changes the subsequent conditional distribution $\pi_{\theta_0}(\cdot\mid \tilde{V},P,\mathcal{E})$, enabling continual adaptation without updating any model parameters. Meanwhile, the frozen learner $\pi_{\theta_0}$ serves as a strong prior that stabilizes the process and mitigates uncontrolled drift caused by experience updates.

\subsection{Anomaly Alignment Reward}
\label{sec:ano_ali_rew}
Anomaly judgment is subjective and scene-dependent, and updating experience solely from absolute supervision signals can be easily biased by incidental preferences.
\revise{Meanwhile, anomaly judgment is also temporally dependent, since the same visual event may lead to different decisions under different preceding and subsequent temporal contexts. Therefore, experience optimization cannot rely only on classification correctness, but should also consider alignment with human risk preferences and reliance on temporal evidence.}
To this end, we design an anomaly alignment reward that evaluates trajectory quality from three aspects: \emph{classification correctness}, \emph{risk-preference alignment}, and \emph{temporal-dependent reasoning}. We use it as the basis for the optimizer to conduct comparative reflection over output groups.

\textbf{Accuracy Reward.}
The accuracy reward provides the most basic directional supervision. We parse an anomaly score $p(o)\in[0,1]$ from an output $o$, and use the video-level weak label $Y\in\{0,1\}$ as supervision:
\begin{equation}
R_{\mathrm{acc}}(o)= Y\cdot p(o) + (1-Y)\cdot (1-p(o)).
\label{eq:r_acc}
\end{equation}
This reward encourages higher anomaly scores when $Y=1$ and lower anomaly scores when $Y=0$.

\textbf{Anomaly Preference Reward.}
Relying only on the accuracy reward may encourage shortcut learning based on spurious correlations in the training data, failing to align with human risk preferences. We therefore introduce an anomaly preference reward. Specifically, we take the scenario experience entry $e_{\mathrm{sce}}^{(j)}$ for the current sample as a positive preference text $e^{+}$, and ask an LLM to generate $H$ semantically similar but preference-mismatched perturbations $\{e^{-}_h\}_{h=1}^{H}$ as hard negatives (e.g., swapping anomaly types, removing key evidence, or altering scene conditions).
\revise{Note that the positive preference text is not free-form model self-feedback; instead, it is constructed from the weak label of the current sample under a rule-based template, so its semantic direction is constrained by external supervisory signals.}
Let $E_t(\cdot)$ denote a text encoder. We encode the output reasoning text and preference rules as $\mathbf{z}(o)=E_t(o)$, $\mathbf{f}^{+}=E_t(e^{+})$, and $\mathbf{f}^{-}_h=E_t(e^{-}_h)$, and compute cosine similarities $s^{+}(o)=\cos(\mathbf{z}(o),\mathbf{f}^{+})$ and $s^{-}_h(o)=\cos(\mathbf{z}(o),\mathbf{f}^{-}_h)$. The preference reward is defined as the softmax probability of the positive preference within the contrastive set:
\begin{equation}
R_{\mathrm{pre}}(o)=
\frac{\exp\!\left(s^{+}(o)/\tau\right)}
{\exp\!\left(s^{+}(o)/\tau\right)+\sum_{h=1}^{H}\exp\!\left(s^{-}_h(o)/\tau\right)},
\label{eq:r_pre}
\end{equation}
where $\tau$ is a temperature hyperparameter. This reward encourages outputs to be semantically consistent with the risk preferences and anomaly cues captured by scenario experience, while staying away from perturbed preferences, thereby improving robustness of preference alignment.

\textbf{Anomaly Temporal Dependency Reward.}
However, preference alignment alone may still lead the model to make anomaly decisions primarily based on static semantics, while neglecting critical temporal evidence. To encourage temporal-dependent reasoning, we introduce an anomaly temporal dependency reward. For the same video, we construct an ordered sampled sequence $\tilde{V}^{\mathrm{ord}}$ and a shuffled sampled sequence $\tilde{V}^{\mathrm{shf}}$, and obtain two output groups $\mathcal{O}^{\mathrm{ord}}$ and $\mathcal{O}^{\mathrm{shf}}$, respectively. \revise{The shuffled sequence uses the same sampled frames as the ordered sequence and only randomly permutes their temporal order. Notably, within LRPO, this ordered/shuffled comparison is used to form a reward signal for experience optimization, rather than to learn temporally sensitive video representations.} Next, we compare the group-wise mean of $R_{\mathrm{acc}}$ and define $\Delta=\bar{R}_{\mathrm{acc}}^{\mathrm{ord}}-\mu\cdot \bar{R}_{\mathrm{acc}}^{\mathrm{shf}}$, where $\mu$ controls the suppression strength for shuffled performance. We only reward improvements where the ordered input outperforms the shuffled one, and apply a soft gate to reduce noise:
\begin{equation}
R_{\mathrm{tem}}^{\mathrm{grp}}=g\cdot \max(0,\Delta),\qquad
g=\mathrm{clip}\!\left(\frac{|\Delta|-\epsilon_0}{\epsilon_1-\epsilon_0},\,0,\,1\right).
\label{eq:r_tem_group}
\end{equation}
Finally, $R_{\mathrm{tem}}^{\mathrm{grp}}$ is assigned only to outputs that are correct under the ordered input (and set to $0$ otherwise), ensuring that the positive advantage is meaningful. This reward measures the stable gain brought by temporal information at the group level, and encourages experience updates to rely on frame-order evidence rather than static semantics.

\subsection{Inference with Experience Selection}
\label{sec:inf_exp_sel}
\textbf{Selector.}
The selector uses a scenario experience retriever $\mathrm{Ret}(\cdot)$ to return the Top-$K$ relevant entries from the scenario experience index set \revise{$\mathcal{K}^{\mathrm{sce}}=\{(e_i^{\mathrm{sce}},\mathbf{k}_{v,i},\mathbf{k}_{t,i})\}$ constructed in \S\ref{sec:two_exp_rep}} for an input video $\tilde{V}$, and concatenates them with generic experience to form the conditional context:
\begin{equation}
\resizebox{\linewidth}{!}{$
\mathrm{Sel}(\mathcal{E},\tilde{V})
=\mathcal{E}^{\mathrm{gen}} \oplus \mathrm{Ret}(\mathcal{K}^{\mathrm{sce}},\tilde{V}),
\qquad \left|\mathrm{Ret}(\mathcal{K}^{\mathrm{sce}},\tilde{V})\right|=K.
$}
\label{eq:experience_injection}
\end{equation}
Here, $\oplus$ denotes concatenation and injection under a predefined itemized format.
Since visual-only retrieval can be confounded by cluttered backgrounds, we complement it with textual semantics that describe the scene and actions. Accordingly, $\mathrm{Ret}(\cdot)$ implements a dual-branch visual-semantic retriever. For a query video $\tilde{V}$, we construct a visual query vector $\mathbf{q}_v=E_v(\tilde{V})$ and a semantic query vector $\mathbf{q}_t$. During inference, we first let the VLM generate a video description $c=\Phi_{\mathrm{VLM}}(\tilde{V}, P_{\mathrm{cap}})$ and then set $\mathbf{q}_t=E_t(c)$; during optimization, $\mathbf{q}_t$ is directly constructed from the scenario experience text associated with the current sample. For any indexed entry $(e_i^{\mathrm{sce}},\mathbf{k}_{v,i},\mathbf{k}_{t,i})\in\mathcal{K}^{\mathrm{sce}}$, we compute cosine similarities $s_{v,i}=\cos(\mathbf{q}_v,\mathbf{k}_{v,i})$ and $s_{t,i}=\cos(\mathbf{q}_t,\mathbf{k}_{t,i})$, normalize the two scores, and fuse them as $\tilde{s}_i=\alpha\,\mathrm{Norm}(s_{v,i})+(1-\alpha)\,\mathrm{Norm}(s_{t,i})$. Finally, $\mathrm{Ret}(\cdot)$ ranks entries by $\tilde{s}_i$ and returns the Top-$K$ scenario experiences.

\textbf{Inference.}
Given a test video $V$, we partition it into segments by sliding along the temporal axis with a fixed stride, where each segment $V_s$ contains $L$ consecutive frames. For each segment $V_s$, the selector constructs an experience context $\mathcal{E}(V_s)=\mathrm{Sel}(\mathcal{E},V_s)$, and we feed $V_s$, the task prompt $P$, and $\mathcal{E}(V_s)$ into the frozen VLM (the same learner VLM used during training) to obtain the anomaly reasoning output $o_s=\Phi_{\mathrm{VLM}}\!\big(V_s, P, \mathcal{E}(V_s)\big)$.

% Experiments
\section{Experiments}
\begin{table*}[t]
\centering
\caption{Comparison with state-of-the-art methods across the XD-Violence, UCF-Crime, and UBnormal datasets.}
\label{tab:sota_frame}
\setlength{\tabcolsep}{4.5pt}
\resizebox{\textwidth}{!}{
\begin{tabular}{llcccccc}
\toprule
\multirow{2}{*}{\textbf{Methods}} &
\multirow{2}{*}{\textbf{Venue}} &
\multirow{2}{*}{\textbf{Supervision}} &
\multirow{2}{*}{\textbf{Explanation}} &
\multirow{1}{*}{\textbf{Training Data}} &
\multicolumn{2}{c}{\textbf{Multi-Scenario}} &
\multicolumn{1}{c}{\textbf{Open-Set}} \\
\cline{5-8}
\rule{0pt}{2.5ex}
& & & & \textbf{XD-Violence / UCF-Crime} &
\textbf{XD-Violence (AP\%)} & \textbf{UCF-Crime (AUC\%)} &
\textbf{UBnormal (AUC\%)} \\
\midrule
\rowcolor{rowgray}
\multicolumn{8}{l}{\textit{\textbf{Tuning-based Methods}}} \\
AED-MAD~\cite{ristea2024self}      & CVPR'24    & Semi                  & \xmark & 100\% / 100\% & --    & --    & 58.50  \\
STPAG~\cite{rai2024video}          & CVPR'24    & Semi                  & \xmark & 100\% / 100\% & --    & --    & 57.98 \\
RFTM~\cite{tian2021weakly}         & ICCV'21    & Weak                  & \xmark & 100\% / 100\% & 77.81 & 84.30 & 64.94 \\
% UR-DMU~\cite{zhou2023dual}         & AAAI'23    & Weak                  & \xmark & 100\% / 100\% & 81.66 & 86.97 & 59.91 \\
PEL4VAD~\cite{pu2024learning}         & TIP'24    & Weak                  & \xmark & 100\% / 100\% & 85.59 & 86.76 & -- \\
VadCLIP~\cite{wu2024vadclip}       & AAAI'24    & Weak                  & \xmark & 100\% / 100\% & 84.51 & 88.02 & --    \\
Ex-VAD~\cite{huangex}              & ICML'25    & Weak                  & \cmark & 100\% / 100\% & 86.52 & 88.29 & --    \\
\midrule
\rowcolor{rowgray}
\multicolumn{8}{l}{\textit{\textbf{Tuning-free Methods}}} \\
ZS CLIP~\cite{radford2021learning} & ICML'21    & Training-free         & \cmark & -- & 17.83 & 53.16 & 46.20  \\
LLaVA-1.5~\cite{liu2024improved}   & CVPR'24    & Training-free         & \cmark & -- & 50.26 & 72.84 & 53.71 \\
LAVAD~\cite{zanella2024harnessing} & CVPR'24    & Training-free         & \cmark & -- & 62.01 & 80.28 & 64.23 \\
AnomalyRuler~\cite{yang2024follow} & ECCV'24    & Training-free         & \cmark & -- & --    & --    & {71.90}  \\
EventVAD~\cite{shao2025eventvad}   & MM'25      & Training-free         & \cmark & -- & 64.04 & 82.03 & --    \\
URF~\cite{lin2026unified}          & NeurIPS'25 & Training-free         & \cmark & -- & 68.07 & 84.28 & 69.02 \\
% PANDA~\cite{yang2026panda}         & NeurIPS'25 & Training-free         & \cmark & -- & 70.16 & 84.89 & 75.78 \\
VADTree~\cite{li2026vadtree}       & NeurIPS'25 & Training-free         & \cmark & -- & 68.85 & 84.74 & --    \\
% VERA~\cite{ye2025vera}             & CVPR'25    & Verbalized Learning   & \cmark & 100\% / 100\% & {70.11} & {86.55} & 71.65    \\
VERA~\cite{ye2025vera}             & CVPR'25    & Verbalized Learning   & \cmark & 100\% / 100\% & {70.11} & {86.55} & 71.65\textsuperscript{*}    \\
% \rowcolor{myblue!50}
\midrule
\multirow{2}{*}{\textbf{LRPO (Ours)}} &
\multirow{2}{*}{--} &
\multirow{2}{*}{Verbalized Learning} &
\cmark & 2.5\% / 6\% & \textbf{73.17} & \textbf{85.36} & \textbf{75.81}\textsuperscript{*} \\
% \rowcolor{myblue!50}
& & &
\cmark & 100\% / 100\% & \textbf{74.09} & \textbf{86.59} & \textbf{76.24}\textsuperscript{*} \\
\bottomrule
\end{tabular}
}
% \vspace{1mm}
% {\scriptsize\noindent\parbox{\textwidth}{%
% \textsuperscript{*} UBnormal results are obtained by directly applying the anomaly reasoning experience learned on XD-Violence.}}
% {\scriptsize\noindent\parbox{\textwidth}{%
% \textsuperscript{*} UBnormal results are obtained by directly applying the anomaly reasoning experience learned on XD-Violence.\\
% \textsuperscript{†} For fair comparison, we directly apply the guiding questions learned by VERA on XD-Violence to infer on UBnormal using the same backbone InternVL3\_5-8B.}}
% {\scriptsize\noindent\parbox{\textwidth}{%
% \textsuperscript{*} We obtain our UBnormal results by directly applying the anomaly reasoning experience learned on XD-Violence, using InternVL3\_5-8B as the backbone.\\
% \textsuperscript{†} We obtain VERA's UBnormal result by directly applying its guiding questions learned on XD-Violence, using the same backbone InternVL3\_5-8B for fair comparison.}}

{\scriptsize\noindent\parbox{\textwidth}{%
\textsuperscript{*} UBnormal results are obtained by directly applying the anomaly reasoning experience (ours) or the guiding questions (VERA) learned on XD-Violence to infer on UBnormal, using InternVL3\_5-8B as the backbone for fair comparison.}}

% \vspace{1mm}
% \vspace{-10px}

\end{table*}

% Ours                               & --        & Verbalized Learning          & \cmark & 6\% & 80.68    & \textbf{75.20} & 74.49 \\

\subsection{Experimental Setup}
\textbf{Datasets.} We conduct experiments on three widely used VAD benchmarks. \textbf{XD-Violence}~\cite{wu2020not} contains 4,754 untrimmed videos spanning 217 hours and covering six types of violence events collected from diverse sources, such as surveillance, movies, and online videos. We use its official split with 3,954 training videos annotated with video-level labels and 800 testing videos annotated with frame-level annotations. \textbf{UCF-Crime}~\cite{sultani2018real} consists of 1,900 untrimmed surveillance videos spanning 128 hours and covering 13 real-world anomaly categories. We follow the standard weakly supervised split, using 1,610 videos for training with video-level labels and 290 videos for testing with frame-level annotations. \textbf{UBnormal}~\cite{acsintoae2022ubnormal} is an open-set virtual dataset generated by Cinema4D, containing 29 scenes with over 236k frames. Following the one-class setting in~\cite{yang2024follow}, we use the same protocol but evaluate only on its test set. \revise{Detailed normal/abnormal frame statistics for the evaluated splits are provided in Appendix~\ref{sec:app_dataset_stats}.}

\textbf{Evaluation Metrics. } We adopt the standard evaluation metrics used in prior work. Specifically, we report the frame-level average precision (AP) on XD-Violence, and the area under the receiver operating characteristic curve (AUC) on UCF-Crime and UBnormal. Compared to AUC, AP is often more suitable for XD-Violence since the dataset is highly imbalanced, and AP places greater emphasis on the positive (violent) class.

\textbf{Implementation Details.}
We use InternVL3\_5-8B~\cite{wang2025internvl3} as the Learner VLM and GPT-OSS-120B~\cite{agarwal2025gpt} as the Optimizer LLM. We train LRPO for $3$ epochs. \revise{For training-subset construction, we fix the annotation budget to $100$ training videos for each dataset, corresponding to $2.5\%$ of XD-Violence and $6\%$ of UCF-Crime due to their different training-set sizes. We randomly sample these videos uniformly across categories to reduce category imbalance.} During training, we sample $M=16$ key frames for each video and draw a group of $G=4$ reasoning trajectories. We cap the size of the learned generic experience repository at $|\mathcal{E}^{\mathrm{gen}}|\le 30$. For experience injection, we use CLIP4CLIP~\cite{luo2022clip4clip} as the text and visual encoders to retrieve the Top-$K$ scenario experiences with $K=10$. During inference, we sparsely sample $L=4$ frames per window with a temporal interval of 16 frames, and slide this window along the video stream. All experiments are conducted on two NVIDIA H800 GPUs. Additional implementation details are provided in Appendix~\ref{sec:app_impl_details}.

\subsection{Main Results}

\textbf{Compare with SOTA Methods.}
Table~\ref{tab:sota_frame} compares LRPO with state-of-the-art VAD methods under both tuning-based and tuning-free settings. The results show that LRPO consistently outperforms existing tuning-free baselines on all three datasets, demonstrating strong effectiveness and competitiveness. Notably, LRPO reaches 73.17\% AP on XD-Violence and 85.36\% AUC on UCF-Crime using only 100 training videos (corresponding to 2.5\% and 6\% of the training set, respectively). 
We observe that using the full training set yields only marginal gains (74.09\% AP on XD-Violence and 86.59\% AUC on UCF-Crime), suggesting that LRPO is highly sample-efficient for experience learning. Rather than fitting the data distribution via parameter updates, LRPO distills linguistically expressed anomaly experiences, making a small, class-balanced set of representative samples sufficient. Moreover, compared to VERA~\cite{ye2025vera}, which optimizes guiding questions to drive binary decisions, LRPO distills reusable anomaly preferences and decision principles, reflecting a more cognition-driven adaptation.

\textbf{Cross-Dataset Transfer and Generalization.}
As shown in Table~\ref{tab:sota_frame}, we evaluate generalization by directly transferring the anomaly reasoning experiences learned by LRPO on XD-Violence to UBnormal, without any additional training. With experience transfer alone, LRPO achieves 76.24\% AUC and sets a new state of the art, suggesting that it distills reusable, language-form anomaly criteria rather than memorizing the source distribution. Under the same backbone and transfer protocol, transferring VERA's guiding questions attains only 71.65\% AUC, indicating that LRPO's anomaly-preference experiences capture more transferable decision principles than question-driven cue focusing.

\begin{figure}[!t]
\vspace{-2pt}
\centerline{\includegraphics[width=\linewidth, keepaspectratio]{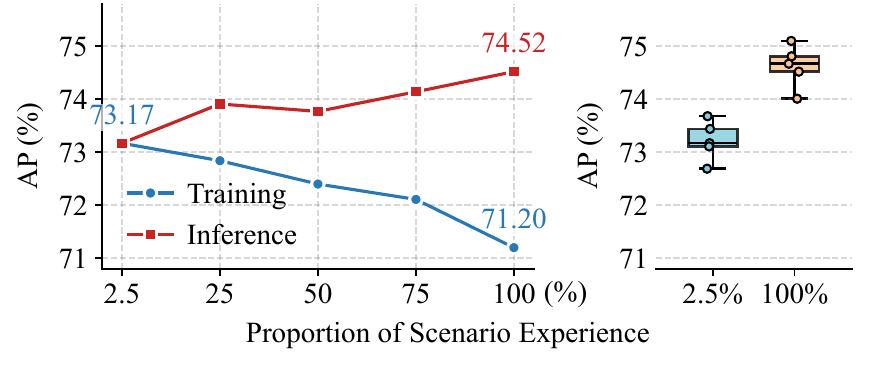}}
% \caption{Analyzing the Influence of Training Video Coverage and Scenario Experience.}
\vspace{-5pt}
% \caption{Scalability and stability of LRPO. (a) The total training data is fixed to 2.5\%: the red curve keeps the general experience fixed and gradually increases the ratio of scenario experience at inference; the blue curve gradually increases the ratio of scenario experience during experience learning. (b) Under random seeds 42-46, we independently sample five 2.5\% training subsets to learn general experiences, and evaluate with 2.5\% and 100\% scenario experience at inference.}
% \caption{Scalability and stability of LRPO. (a) Scaling scenario experience at inference improves performance, whereas increasing scenario experience during experience learning degrades it. (b) LRPO remains stable across five random 2.5\% training-subset samples.}
\caption{Scalability and stability of LRPO. (a) Scaling scenario experience at inference improves performance, whereas scaling it during experience learning degrades it. (b) LRPO remains stable across five random 2.5\% training-subset samples.}
\label{fig:trend_stability}
% \vspace{-8pt}
\end{figure}

\textbf{Scalability and Stability Analysis.}
As shown in Figure~\ref{fig:trend_stability}(a), with the general experience learned from 2.5\% training data fixed, increasing the ratio of scenario experience at inference (red curve) consistently improves performance (73.17\% to 74.52\% AP), indicating that LRPO continues to benefit from scaling scenario experience. In contrast, scaling scenario experience during experience learning (blue curve) degrades performance (73.17\% to 71.20\% AP). We attribute this to shortcut learning: excessive scenario experience encourages reliance on scenario-specific cues, weakening the induction of reusable general experience and harming robustness on rare cases where matching scenario experiences are hard to retrieve. \revise{This highlights the importance of learning high-quality experience rather than simply accumulating more scenario-specific cues.}
To further assess stability, we independently sample five 2.5\% training subsets with random seeds 42-46 to learn general experiences, and evaluate with 2.5\% and 100\% scenario experience at inference. As shown in Figure~\ref{fig:trend_stability}(b), the results are stable at 73.17\% $\pm$ 0.51 and 74.52\% $\pm$ 0.58, respectively, indicating that LRPO is robust to training-subset sampling.

\iffalse
\caption{LRPO 的可扩展性与稳定性。(a) 总训练数据固定为 2.5\%：红线固定通用经验不变，在推理阶段逐步增加场景经验比例；蓝线在经验学习阶段逐步增加场景经验比例。 (b) 在随机种子 42-46 下独立随机抽取 5 份 2.5\% 训练子集以学习通用经验，并在推理阶段分别使用 2.5\% 与 100\% 的场景经验进行评估。}

\caption{LRPO 的可扩展性与稳定性。(a) 总训练数据固定为 2.5\%：红线固定通用经验不变，在推理阶段逐步增加场景经验比例；蓝线在通用经验学习阶段逐步增加场景经验比例。 (b) 在随机种子 42-46 下重复 5 次从 2.5\% 训练数据采样学习通用经验，并在推理阶段分别使用 2.5\% 与 100\% 的场景经验进行评估。}

\caption{LRPO 的可扩展性与稳定性。(a) 总训练数据固定为 2.5\%：红线固定由 2.5\% 数据学习的通用经验不变，在推理阶段逐步增加场景经验比例；蓝线在经验学习阶段改变 2.5\% 数据中场景经验的占比（其余用于通用经验）。(b) 在随机种子 42-46 下重复 5 次从 2.5\% 训练数据随机采样学习通用经验，并在推理阶段分别使用 2.5\% 与 100\% 的场景经验进行评估。}

在随机种子 42-46 下从完整训练集中独立随机抽取 2.5\% 的训练子集各一次（共 5 次），并基于每个子集学习通用经验

\caption{LRPO 的可扩展性与稳定性。(a) 固定由 2.5\% 训练数据学习得到的通用经验不变，在推理阶段逐步增加场景经验比例（红线）；对比在经验学习阶段使用不同比例的场景经验进行学习（蓝线）。(b) 在随机种子 42-46 下重复 5 次从 2.5\% 训练数据随机采样学习通用经验，并在推理阶段分别使用 2.5\% 与 100\% 的场景经验进行评估。}

\caption{LRPO 的可扩展性与稳定性。(a) 固定 2.5\% 训练数据学习通用经验，在推理（红线）和训练（蓝线）阶段逐步增加场景经验比例（红线）。(b) 在随机种子 42-46 下，随机 5 次采样 2.5\% 训练数据学习通用经验，并在推理阶段分别使用 2.5\% 与 100\% 的场景经验。}

\caption{LRPO 的可扩展性与稳定性。(a) 固定由 2.5\% 训练数据学习的通用经验不变，推理阶段逐步增加场景经验比例带来持续增益（红线）；相反，若在经验学习阶段使用更高比例的场景经验进行训练，则整体性能下降（蓝线）。(b) 在随机种子 42-46 下重复 5 次从 2.5\% 训练数据采样学习通用经验，并在推理时分别使用 2.5\% 与 100\% 场景经验进行评估，结果表现稳定。}

\caption{LRPO 的可扩展性与稳定性。(a) 固定 2.5\% 数据学习的通用经验，推理阶段增加场景经验可提升性能（红线），而训练阶段增加场景经验会降低性能（蓝线）。(b) 随机种子 42-46 下重复 5 次采样学习通用经验，在 2.5\% 与 100\% 场景经验推理设置下结果稳定。}

\caption{LRPO 的可扩展性与稳定性分析。(a) 可扩展性：固定由 2.5\% 训练数据学习得到的通用经验，在推理阶段逐步增加场景经验比例可持续提升性能（红线）；而在经验学习阶段增加场景经验比例则导致性能下降（蓝线）。(b) 稳定性：在随机种子 42-46 下重复 5 次从 2.5\% 训练数据采样学习通用经验，并分别在推理时使用 2.5\% 与 100\% 的场景经验进行评估。}

\fi

\textbf{Effectiveness across Model Families and Scales.}
Table~\ref{tab:abla_vlm_llm} evaluates LRPO across different model families and parameter scales. LRPO delivers consistent gains with both LLaVA-NeXT-7B~\cite{liu2024improved} and InternVL3\_5-8B~\cite{wang2025internvl3} as the learner (13.83\% and 13.24\% AP improvements), demonstrating strong cross-backbone applicability. The choice of optimizer is also flexible: using InternLM3-8B~\cite{cai2024internlm2} improves InternVL3\_5-8B from 59.93\% AP to 72.77\% AP, while a stronger GPT-OSS-120B~\cite{agarwal2025gpt} further reaches 73.17\% AP, indicating that more capable reflection and experience optimization can yield additional gains. \revise{Meanwhile, this also indicates that LRPO does not rely on the strongest external model to remain effective.}

\begin{table}[!ht]
% \vspace{-5pt}
\centering
\caption{Effectiveness of the proposed LRPO under different learner-optimizer pairs. %LRPO improves the learner by injecting optimized linguistic experiences.
}
\label{tab:abla_vlm_llm}
\resizebox{\linewidth}{!}{
\begin{tabular}{l l l l}
\toprule
\textbf{Learner} & \textbf{Optimizer} & \textbf{Method} & \textbf{AP (\%)} \\
\midrule
\multirow{2}{*}{LLaVA-NeXT-7B}
 & -- & Baseline & 32.94 \small{\textcolor{white}{(+0.000)}} \\
 & GPT-OSS-120B & LRPO (Ours) & 46.77 \small{\textcolor{black}{(+13.83)}} \\
\midrule
\multirow{3}{*}{InternVL3\_5-8B}
 & -- & Baseline & 59.93 \small{\textcolor{white}{(+0.000)}} \\
 & InternLM3-8B & LRPO (Ours) & 72.77 \small{\textcolor{black}{(+12.84)}} \\
 & GPT-OSS-120B & LRPO (Ours) & \textbf{73.17} \small{\textcolor{black}{(+13.24)}} \\
\bottomrule
\end{tabular}
}
\end{table}

\subsection{Ablation Study}

\textbf{Effect of General and Scenario Experiences.}
As shown in Table~\ref{tab:abla_overall}, we analyze the effects of general experience $\mathcal{E}^{\mathrm{gen}}$ and scenario experience $\mathcal{E}^{\mathrm{sce}}$. The learner achieves 59.93\% AP without experience injection. Injecting $\mathcal{E}^{\mathrm{gen}}$ and learning with the accuracy reward $R_{\mathrm{acc}}$ improves performance to 68.48\% AP, suggesting that LRPO distills general experience that effectively strengthens the learner's anomaly-aware cognition. Further adding $\mathcal{E}^{\mathrm{sce}}$ under the same setting increases performance to 70.58\% AP, indicating that scenario experience complements $\mathcal{E}^{\mathrm{gen}}$ by calibrating decisions with context-specific cues.

\begin{table}[!htp]
% \vspace{-5pt}
\caption{Ablation study on XD-Violence.}
\centering
% \scriptsize 
\resizebox{0.85\linewidth}{!}{
\begin{tabular}{cc|ccc|l}
\toprule
\multicolumn{2}{c|}{\textbf{Experience}} &
\multicolumn{3}{c|}{\textbf{Reward}} &
\multicolumn{1}{c}{\multirow{2}{*}{\textbf{AP(\%)}}} \\
\cline{1-5}
\rule{0pt}{2.5ex}
$\mathcal{E}^{\mathrm{gen}}$& $\mathcal{E}^{\mathrm{sce}}$&
$R_{\mathrm{acc}}$& $R_{\mathrm{pre}}$& $R_{\mathrm{tem}}$&
\\
\midrule
% \xmark & \xmark & \xmark & \xmark & \xmark & 59.93 \small{\textcolor{white}{(+0.00)}} \\
% \cmark & \xmark & \xmark & \xmark & \xmark & 67.88 \small{\textcolor{gray}{(+7.95)}} \\
% \cmark & \cmark & \xmark & \xmark & \xmark & 71.71 \small{\textcolor{gray}{(+11.78)}} \\
% \cmark & \xmark & \cmark & \xmark & \xmark & 00.00 \small{\textcolor{gray}{(+2.98)}} \\
% \cmark & \xmark & \xmark & \cmark & \xmark & 00.00 \small{\textcolor{gray}{(+0.69)}} \\
% \cmark & \xmark & \xmark & \xmark & \cmark & 00.00 \small{\textcolor{gray}{(+1.66)}} \\
% \cmark & \xmark & \cmark & \cmark & \cmark & 72.11 \small{\textcolor{gray}{(+12.18)}} \\

\xmark & \xmark & \xmark & \xmark & \xmark & 59.93 \small{\textcolor{white}{(+0.00)}} \\
\cmark & \xmark & \cmark & \xmark & \xmark & 68.48 \small{\textcolor{gray}{(+8.55)}} \\
\cmark & \cmark & \cmark & \xmark & \xmark & 70.58 \small{\textcolor{gray}{(+10.65)}} \\
\cmark & \xmark & \cmark & \cmark & \xmark & 69.78 \small{\textcolor{gray}{(+9.85)}} \\
\cmark & \xmark & \cmark & \xmark & \cmark & 70.06 \small{\textcolor{gray}{(+10.13)}} \\
\cmark & \xmark & \cmark & \cmark & \cmark & 71.91 \small{\textcolor{gray}{(+11.98)}} \\

% \rowcolor{myblue!50}
\cmark & \cmark & \cmark & \cmark & \cmark & \textbf{73.17} \small{(+13.24)} \\
\bottomrule
\end{tabular}
}
\label{tab:abla_overall}
\end{table}

\textbf{Effect of Reward Components.}
Table~\ref{tab:abla_overall} examines how the anomaly alignment reward affects experience learning. With experience fixed to $\mathcal{E}^{\mathrm{gen}}$, using only $R_{\mathrm{acc}}$ achieves 68.48\% AP. Adding anomaly preference reward $R_{\mathrm{pre}}$ improves to 69.78\% AP, indicating that it discourages shortcut reasoning inconsistent with the desired anomaly preference (risk bias). Adding anomaly temporal dependency $R_{\mathrm{tem}}$ improves to 70.06\% AP, suggesting that temporal-consistency constraints promote temporally grounded reasoning and yield more reliable experiences. Combining $R_{\mathrm{acc}}$, $R_{\mathrm{pre}}$, and $R_{\mathrm{tem}}$ further reaches 71.91\% AP, demonstrating their complementarity. Finally, injecting $\mathcal{E}^{\mathrm{sce}}$ on top of the full reward design achieves 73.17\% AP, showing that reward constraints and scenario-experience injection jointly enhance the quality and utility of learned linguistic experiences.

\revise{\textbf{Effect of Experience Optimization.}}
\revise{To better separate the effect of optimization from that of richer textual experience augmentation, we keep the experience injection format fixed and examine performance as the experience is progressively refined during LRPO. As shown in Table~\ref{tab:expe_optimization_progress}, AP improves steadily from 67.63\% to 73.17\% across the initial, one-third, two-thirds, and final stages. This suggests that the gain comes not only from adding experience, but also from continually improving its quality through optimization.}
\begin{table}[!ht]
\centering
\caption{\protect\revise{Effect of experience optimization during LRPO.}}
\label{tab:expe_optimization_progress}
\resizebox{0.85\linewidth}{!}{
\begin{tabular}{lcccc}
\toprule
\revise{\textbf{Experience stage}} &
\revise{\textbf{Initial}} &
\revise{\textbf{1/3}} &
\revise{\textbf{2/3}} &
\revise{\textbf{Final}} \\
\midrule
\revise{\textbf{AP (\%)}} &
\revise{67.63} &
\revise{70.04} &
\revise{72.96} &
\revise{\textbf{73.17}} \\
\bottomrule
\end{tabular}
}
\end{table}

\begin{table}[t]
\centering
\caption{Ablation on scenario experience injection strategies.}% on the XD-Violence dataset.}
% \vspace{2pt}
\label{tab:tab_abla_expe_inje}
\resizebox{\linewidth}{!}{
\begin{tabular}{l l}
\toprule
\textbf{Scenario experience injection strategy} & \textbf{AP (\%)} \\
\midrule
No Injection & 71.91 \small{\textcolor{white}{(+0.00)}}\\
Random Injection & 71.61 \small{\textcolor{gray}{(-0.30)}}\\
Visual-Retrieved Injection & 72.54 \small{\textcolor{gray}{(+0.63)}} \\
\textbf{Visual-Semantic Retrieved Injection (Ours)} & \textbf{73.17} \small{\textcolor{black}{(+1.26)}} \\
\bottomrule
\end{tabular}
}
\end{table}

\textbf{Ablation of Scenario Experience Injection Strategy.}
Keeping other settings unchanged, Table~\ref{tab:tab_abla_expe_inje} ablates scenario experience injection strategies. Without injection, the model reaches 71.91\% AP, while random injection drops to 71.61\% AP due to noisy or irrelevant experiences. Visual retrieval improves to 72.54\% AP, indicating that retrieval selects scenario-matched experiences for effective contextual constraints. Our visual-semantic retrieval achieves the best 73.17\% AP, showing that jointly modeling visual similarity and semantic matching yields more reliable scenario-boundary localization and experience utilization.

\revise{\textbf{Sensitivity to Generator Quality.}}
\revise{We further examine whether the preference reward is sensitive to generator quality. As shown in Tables~\ref{tab:sce_generator_sensitivity} and~\ref{tab:neg_generator_sensitivity}, replacing InternVL3\_5-38B with weaker InternVL3\_5-14B/8B for scenario experience construction only slightly reduces AP from 73.17\% to 73.02\%/72.59\%. Similarly, using GPT-OSS-20B instead of GPT-OSS-120B for negative sample generation still achieves 72.28\% AP. These performance drops are smaller than the gains brought by the corresponding components in Table~\ref{tab:abla_overall}, indicating that generator quality mainly affects the performance ceiling, while the main improvements come from the core design of LRPO.}

\begin{table}[!ht]
\centering
\caption{\protect\revise{Sensitivity of $R_{\mathrm{pre}}$ to scenario experience generator scale using InternVL3\_5~\cite{wang2025internvl3}.}}
\label{tab:sce_generator_sensitivity}
\resizebox{\linewidth}{!}{
\begin{tabular}{lccc}
\toprule
\revise{\textbf{Scenario experience generator scale}} &
\revise{\textbf{8B}} &
\revise{\textbf{14B}} &
\revise{\textbf{38B}} \\
\midrule
\revise{\textbf{AP (\%)}} & \revise{72.59} & \revise{73.02} & \revise{\textbf{73.17}} \\
\bottomrule
\end{tabular}
}
\end{table}

\begin{table}[!ht]
\centering
\caption{\protect\revise{Sensitivity of $R_{\mathrm{pre}}$ to negative sample generator scale using GPT-OSS~\cite{agarwal2025gpt}.}}
\label{tab:neg_generator_sensitivity}
\resizebox{0.8\linewidth}{!}{
\begin{tabular}{lcc}
\toprule
\revise{\textbf{Negative sample generator scale}} &
\revise{\textbf{20B}} &
\revise{\textbf{120B}} \\
\midrule
\revise{\textbf{AP (\%)}} & \revise{72.28} & \revise{\textbf{73.17}} \\
\bottomrule
\end{tabular}
}
\end{table}

\begin{figure}[!htb]
% \vspace{-5pt}
\centerline{\includegraphics[width=\linewidth, keepaspectratio]{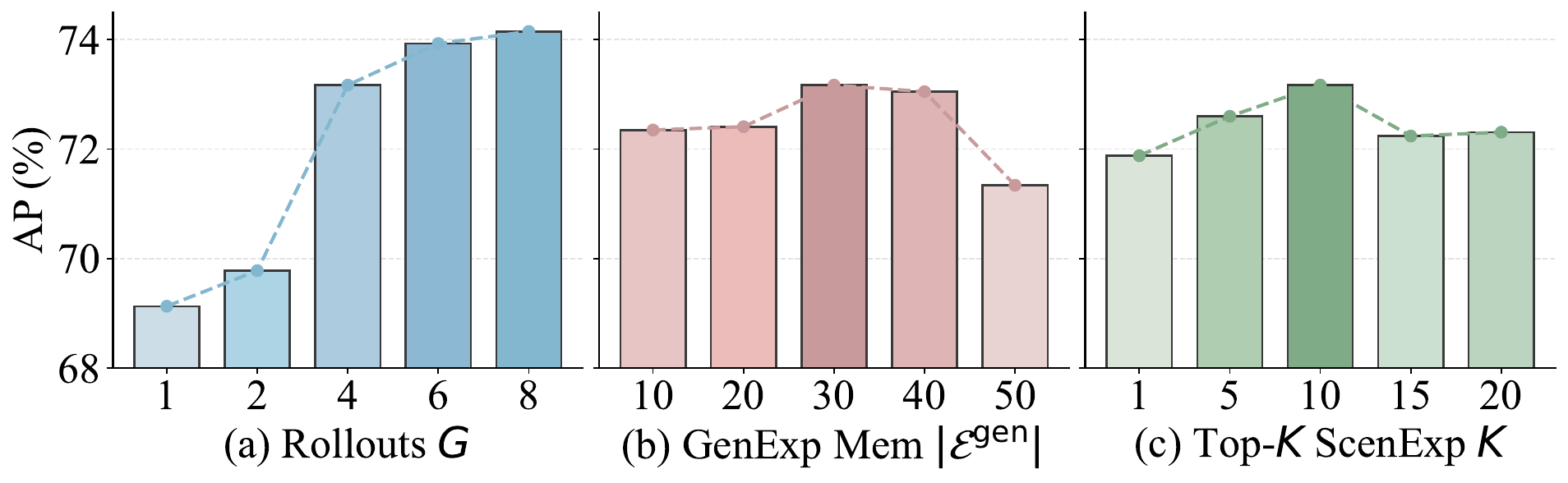}}
% \vspace{-5pt}
\caption{Hyperparameter ablation of LRPO. We report AP (\%) when varying (a) rollout group size $G$, (b) general experience memory size $|\mathcal{E}^{\mathrm{gen}}|$, and (c) the number of retrieved scenario experiences (Top-$K$) $K$.}

\label{fig:para_abla}
\end{figure}

\textbf{Sensitivity to Hyperparameters.}
As shown in Figure~\ref{fig:para_abla}, we analyze LRPO's sensitivity to key hyperparameters. (a) Increasing the rollout group size $G$ improves performance and saturates at larger $G$, indicating that more trajectories enable more stable estimation of group-relative semantic advantages and better experience distillation. (b) The general experience memory size $|\mathcal{E}^{\mathrm{gen}}|$ exhibits a ``moderate-is-best'' trend: too small a memory limits coverage, whereas too large a memory may introduce redundancy and noise. (c) The Top-$K$ used for scenario experience retrieval also presents a trade-off: moderate $K$ is beneficial, while overly large $K$ leads to saturated or slightly degraded performance due to increased noise and interference. Overall, LRPO remains stable across a wide range of hyperparameters, demonstrating strong robustness.

\begin{figure}[!htb]
\centerline{\includegraphics[width=1.0\linewidth, keepaspectratio]{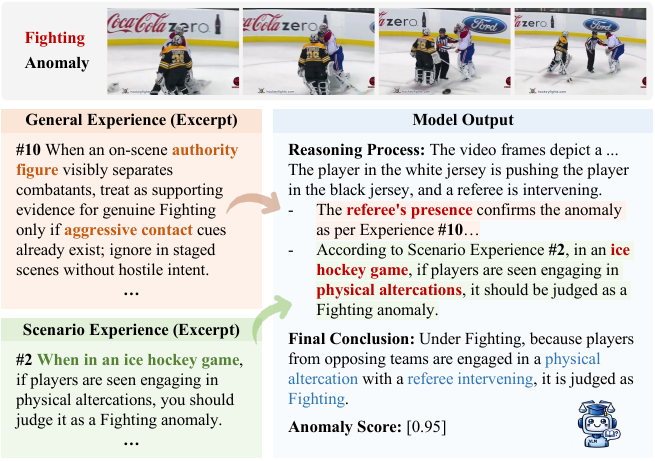}}
% \caption{Qualitative analysis.}
% \caption{Qualitative example of LRPO, showing how general experience and scenario experience jointly support interpretable anomaly reasoning.}
% \caption{Qualitative example of LRPO, showing how general and scenario experiences jointly align anomaly reasoning with risk preferences.}
% \caption{Qualitative example on video ``v=251\_\_\_mEwZA'' from XD-Violence, showing how general and scenario experiences jointly align anomaly reasoning with risk preferences.}
% \caption{Qualitative example on video ``v=251\_\_\_mEwZA'', showing how LRPO leverages general and scenario experiences to produce anomaly-preference-aligned reasoning.}
\caption{Qualitative example on video ``v=251\_\_\_mEwZA'': by injecting learned general experience and retrieved scenario experience, the model produces anomaly preference aligned reasoning.}
% \caption{Qualitative example on video ``v=251\_\_\_mEwZA'' where injecting learned general experience and retrieved scenario experience leads to reasoning aligned with anomaly preferences.}
\label{fig:vis_qual}
% \vspace{-8pt}
\end{figure}

\subsection{Qualitative Analysis}
Figure~\ref{fig:vis_qual} presents a fighting anomaly example to illustrate how LRPO performs interpretable anomaly reasoning by combining shared general experiences with sample-specific retrieved scenario experiences. For clarity, we show one cited general experience and one retrieved scenario experience (more experience entries are provided in Appendix~\ref{sec:app_expe}). The model cites a general experience (e.g., \#10) that treats the authority figure as auxiliary evidence and requires aggressive-contact cues to confirm genuine fighting, and retrieves an ice-hockey scenario experience (e.g., \#2) to calibrate the decision boundary. This example demonstrates that LRPO grounds predictions in human-readable experiences and aligns reasoning with human risk preferences, improving interpretability and decision consistency.

\begin{figure}[!htb]
\centerline{\includegraphics[width=1.0\linewidth, keepaspectratio]{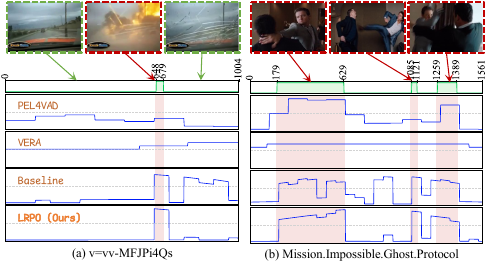}}
\caption{Visualization of anomaly detection results. The gray horizontal line at 0.5 indicates the anomaly detection threshold.}
\label{fig:vis_score}
% \vspace{-8pt}
\end{figure}

We visualize per-frame anomaly scores on two videos in Figure~\ref{fig:vis_score}, comparing a tuning-based method (PEL4VAD~\cite{pu2024learning}), a tuning-free method (VERA~\cite{ye2025vera}), the experience-free Baseline, and LRPO. The Baseline exhibits large score fluctuations and often responds strongly on non-anomalous segments, while VERA produces smoother curves but with limited discriminability. PEL4VAD captures some anomalous events yet shows shifted boundaries or fragmented peaks. In contrast, LRPO concentrates high scores on true anomalous intervals while keeping normal segments low, indicating more consistent temporal localization and fewer false alarms. Additional qualitative analyses and visualizations are provided in the appendix~\ref{sec:app_qual}.

% Conclusion
\section{Limitations}
\revise{Although LRPO can progressively learn and refine anomaly experience without manually enumerating all possible anomaly rules, the coverage of the learned repository remains dependent on the diversity of available training data. Rare scenes or unseen anomaly mechanisms absent from the training subset may therefore be insufficiently represented, motivating future work on automatic experience expansion and validation for open-world anomaly scenarios.}

\section{Conclusion}
We propose LRPO, a novel tuning-free framework for VAD that distills group-relative semantic advantages from multiple reasoning trajectories into a linguistically expressed anomaly experience prior. To support both transferability and contextual calibration, LRPO constructs general and scenario experiences, capturing transferable anomaly preferences and context-dependent anomaly rules, respectively. We further introduce an anomaly alignment reward to optimize experiences, encouraging consistency with human risk preferences while reinforcing temporally grounded reasoning. Extensive experimental results validate the effectiveness and strong competitiveness of LRPO.

\section*{Impact Statement}
% This paper presents work whose goal is to advance the field of 
% Machine Learning. There are many potential societal consequences 
% of our work, none which we feel must be specifically highlighted here.

The proposed LRPO framework can support video anomaly detection in safety-critical scenarios by improving the localization and interpretation of abnormal events, such as violence, accidents, or other emergency situations. Its tuning-free and language-editable experience mechanism may also reduce the cost of adapting vision-language models to new surveillance environments.

Meanwhile, video anomaly detection systems may raise privacy, surveillance, and false-alarm concerns if deployed without appropriate safeguards. Responsible use requires lawful data collection, privacy protection, careful evaluation across diverse scenarios, and human oversight in high-stakes decision-making.
% Authors are \textbf{required} to include a statement of the potential 
% broader impact of their work, including its ethical aspects and future 
% societal consequences. This statement should be in an unnumbered 
% section at the end of the paper (co-located with Acknowledgements -- 
% the two may appear in either order, but both must be before References), 
% and does not count toward the paper page limit. In many cases, where 
% the ethical impacts and expected societal implications are those that 
% are well established when advancing the field of Machine Learning, 
% substantial discussion is not required, and a simple statement such 
% as the following will suffice:

% ``This paper presents work whose goal is to advance the field of 
% Machine Learning. There are many potential societal consequences 
% of our work, none which we feel must be specifically highlighted here.''

% The above statement can be used verbatim in such cases, but we 
% encourage authors to think about whether there is content which does 
% warrant further discussion, as this statement will be apparent if the 
% paper is later flagged for ethics review.

\section*{Acknowledgements}
This work was supported in part by the New Generation Artificial Intelligence-National Science and Technology Major Project under Grant No. 2025ZD0123601, in part by the National Natural Science Foundation of China under Grants No. 62472060 and 62221005, in part by the Science and Technology Innovation Key R\&D Program of Chongqing under Grant No. CSTB2023TIAD-STX0016, in part by the Natural Science Foundation of Chongqing under Grants No. CSTB2024NSCQ-QCXMX0060, in part by the China Postdoctoral Science Foundation under Grant No. 2025MD774186, in part by the Chongqing Special Postdoctoral Research Funding under Grant No. 2024CQBSHTB2002, and in part by the Chongqing University of Posts and Telecommunications Ph.D. Innovative Talents Project under Grant No. BYJS202404.

% Uncomment to include figures and tables (examples)
% \input{figures/fig1_overview}
% \input{figures/fig2_performance}
% \input{tables/table1_main_results}
% \input{tables/table2_ablation}
% \input{tables/table3_efficiency}

\nocite{langley00}

\bibliography{bib/example_paper}

@inproceedings{langley00,
 author    = {P. Langley},
 title     = {Crafting Papers on Machine Learning},
 year      = {2000},
 pages     = {1207--1216},
 editor    = {Pat Langley},
 booktitle     = {Proceedings of the 17th International Conference
              on Machine Learning (ICML 2000)},
 address   = {Stanford, CA},
 publisher = {Morgan Kaufmann}
}

@inproceedings{zhang2025holmes,
  title={Holmes-vau: Towards long-term video anomaly understanding at any granularity},
  author={Zhang, Huaxin and Xu, Xiaohao and Wang, Xiang and Zuo, Jialong and Huang, Xiaonan and Gao, Changxin and Zhang, Shanjun and Yu, Li and Sang, Nong},
  booktitle={Proceedings of the computer vision and pattern recognition conference},
  pages={13843--13853},
  year={2025}
}

@inproceedings{cai2025hiprobe,
  title={Hiprobe-vad: Video anomaly detection via hidden states probing in tuning-free multimodal llms},
  author={Cai, Zhaolin and Li, Fan and Zheng, Ziwei and Qin, Yanjun},
  booktitle={Proceedings of the 33rd ACM International Conference on Multimedia},
  pages={592--601},
  year={2025}
}

@inproceedings{wu2020not,
  title={Not only look, but also listen: Learning multimodal violence detection under weak supervision},
  author={Wu, Peng and Liu, Jing and Shi, Yujia and Sun, Yujia and Shao, Fangtao and Wu, Zhaoyang and Yang, Zhiwei},
  booktitle={European conference on computer vision},
  pages={322--339},
  year={2020},
  organization={Springer}
}

@inproceedings{du2024uncovering,
  title={Uncovering what why and how: A comprehensive benchmark for causation understanding of video anomaly},
  author={Du, Hang and Zhang, Sicheng and Xie, Binzhu and Nan, Guoshun and Zhang, Jiayang and Xu, Junrui and Liu, Hangyu and Leng, Sicong and Liu, Jiangming and Fan, Hehe and others},
  booktitle={Proceedings of the IEEE/CVF Conference on Computer Vision and Pattern Recognition},
  pages={18793--18803},
  year={2024}
}

@inproceedings{zhang2024video,
  title={Video Anomaly Detection via Progressive Learning of Multiple Proxy Tasks},
  author={Zhang, Menghao and Wang, Jingyu and Qi, Qi and Ren, Pengfei and Sun, Haifeng and Zhuang, Zirui and Wang, Huazheng and Zhang, Lei and Liao, Jianxin},
  booktitle={Proceedings of the 32nd ACM International Conference on Multimedia},
  pages={4719--4728},
  year={2024}
}

@article{zhu2024advancing,
  title={Advancing video anomaly detection: A concise review and a new dataset},
  author={Zhu, Liyun and Wang, Lei and Raj, Arjun and Gedeon, Tom and Chen, Chen},
  journal={Advances in Neural Information Processing Systems},
  volume={37},
  pages={89943--89977},
  year={2024}
}

@inproceedings{zhang2024multi,
  title={Multi-scale video anomaly detection by multi-grained spatio-temporal representation learning},
  author={Zhang, Menghao and Wang, Jingyu and Qi, Qi and Sun, Haifeng and Zhuang, Zirui and Ren, Pengfei and Ma, Ruilong and Liao, Jianxin},
  booktitle={Proceedings of the IEEE/CVF conference on computer vision and pattern recognition},
  pages={17385--17394},
  year={2024}
}

@inproceedings{yan2023feature,
  title={Feature prediction diffusion model for video anomaly detection},
  author={Yan, Cheng and Zhang, Shiyu and Liu, Yang and Pang, Guansong and Wang, Wenjun},
  booktitle={Proceedings of the IEEE/CVF international conference on computer vision},
  pages={5527--5537},
  year={2023}
}

@article{cao2024context,
  title={Context recovery and knowledge retrieval: A novel two-stream framework for video anomaly detection},
  author={Cao, Congqi and Lu, Yue and Zhang, Yanning},
  journal={IEEE Transactions on Image Processing},
  volume={33},
  pages={1810--1825},
  year={2024},
  publisher={IEEE}
}

@article{yang2026panda,
  title={Panda: Towards generalist video anomaly detection via agentic ai engineer},
  author={Yang, Zhiwei and Gao, Chen and Shou, Mike Zheng},
  journal={Advances in Neural Information Processing Systems},
  volume={38},
  pages={83182--83211},
  year={2026}
}

@inproceedings{sultani2018real,
  title={Real-world anomaly detection in surveillance videos},
  author={Sultani, Waqas and Chen, Chen and Shah, Mubarak},
  booktitle={Proceedings of the IEEE conference on computer vision and pattern recognition},
  pages={6479--6488},
  year={2018}
}

@inproceedings{zanella2024harnessing,
  title={Harnessing large language models for training-free video anomaly detection},
  author={Zanella, Luca and Menapace, Willi and Mancini, Massimiliano and Wang, Yiming and Ricci, Elisa},
  booktitle={Proceedings of the IEEE/CVF Conference on Computer Vision and Pattern Recognition},
  pages={18527--18536},
  year={2024}
}

@article{pu2024learning,
  title={Learning prompt-enhanced context features for weakly-supervised video anomaly detection},
  author={Pu, Yujiang and Wu, Xiaoyu and Yang, Lulu and Wang, Shengjin},
  journal={IEEE Transactions on Image Processing},
  volume={33},
  pages={4923--4936},
  year={2024},
  publisher={IEEE}
}

@article{leng2024beyond,
  title={Beyond euclidean: Dual-space representation learning for weakly supervised video violence detection},
  author={Leng, Jiaxu and Wu, Zhanjie and Tan, Mingpi and Liu, Yiran and Gan, Ji and Chen, Haosheng and Gao, Xinbo},
  journal={Advances in Neural Information Processing Systems},
  volume={37},
  pages={17373--17397},
  year={2024}
}

@article{leng2025piercingeye,
  title={Piercingeye: Dual-space video violence detection with hyperbolic vision-language guidance},
  author={Leng, Jiaxu and Wu, Zhanjie and Tan, Mingpi and Mo, Mengjingcheng and Zheng, Jiankang and Li, Qingqing and Gan, Ji and Gao, Xinbo},
  journal={IEEE Transactions on Pattern Analysis and Machine Intelligence},
  year={2025},
  publisher={IEEE}
}

@inproceedings{yu2022modality,
  title={Modality-aware contrastive instance learning with self-distillation for weakly-supervised audio-visual violence detection},
  author={Yu, Jiashuo and Liu, Jinyu and Cheng, Ying and Feng, Rui and Zhang, Yuejie},
  booktitle={Proceedings of the 30th ACM international conference on multimedia},
  pages={6278--6287},
  year={2022}
}

@article{zhou2024learning,
  title={Learning weakly supervised audio-visual violence detection in hyperbolic space},
  author={Zhou, Xiao and Peng, Xiaogang and Wen, Hao and Luo, Yikai and Yu, Keyang and Yang, Ping and Wu, Zizhao},
  journal={Image and Vision Computing},
  volume={151},
  pages={105286},
  year={2024},
  publisher={Elsevier}
}

@article{huang2026vad,
  title={Vad-r1: Towards video anomaly reasoning via perception-to-cognition chain-of-thought},
  author={Huang, Chao and Wang, Benfeng and Wang, Wei and Wen, Jie and Liu, Chengliang and Shen, Li and Cao, Xiaochun},
  journal={Advances in neural information processing systems},
  volume={38},
  pages={118486--118518},
  year={2026}
}

@article{mo2026a2seek,
  title={A2seek: Towards reasoning-centric benchmark for aerial anomaly understanding},
  author={Mo, Mengjingcheng and Tong, Xinyang and Tan, Mingpi and Leng, Jiaxu and Zheng, Jiankang and Liu, Yiran and Chen, Haosheng and Gan, Ji and Li, Weisheng and Gao, Xinbo},
  journal={Advances in Neural Information Processing Systems},
  volume={38},
  year={2026}
}

@inproceedings{mo2024nexusad,
  title={Nexusad: Exploring the nexus for multimodal perception and comprehension of corner cases in autonomous driving},
  author={Mo, Mengjingcheng and Wang, Jingxin and Wang, Like and Chen, Haosheng and Gu, Changjun and Leng, Jiaxu and Gao, Xinbo},
  booktitle={ECCV 2024 Workshop on Multimodal Perception and Comprehension of Corner Cases in Autonomous Driving},
  year={2024}
}

@article{luo2022clip4clip,
  title={Clip4clip: An empirical study of clip for end to end video clip retrieval and captioning},
  author={Luo, Huaishao and Ji, Lei and Zhong, Ming and Chen, Yang and Lei, Wen and Duan, Nan and Li, Tianrui},
  journal={Neurocomputing},
  volume={508},
  pages={293--304},
  year={2022},
  publisher={Elsevier}
}

@inproceedings{acsintoae2022ubnormal,
  title={{Ubnormal}: New benchmark for supervised open-set video anomaly detection},
  author={Acsintoae, Andra and Florescu, Andrei and Georgescu, Mariana-Iuliana and Mare, Tudor and Sumedrea, Paul and Ionescu, Radu Tudor and Khan, Fahad Shahbaz and Shah, Mubarak},
  booktitle={Proceedings of the IEEE/CVF conference on computer vision and pattern recognition},
  pages={20143--20153},
  year={2022}
}

@article{shao2024deepseekmath,
  title={Deepseekmath: Pushing the limits of mathematical reasoning in open language models},
  author={Shao, Zhihong and Wang, Peiyi and Zhu, Qihao and Xu, Runxin and Song, Junxiao and Bi, Xiao and Zhang, Haowei and Zhang, Mingchuan and Li, YK and Wu, Yang and others},
  journal={arXiv preprint arXiv:2402.03300},
  year={2024}
}

@article{brown2020language,
  title={Language models are few-shot learners},
  author={Brown, Tom and Mann, Benjamin and Ryder, Nick and Subbiah, Melanie and Kaplan, Jared D and Dhariwal, Prafulla and Neelakantan, Arvind and Shyam, Pranav and Sastry, Girish and Askell, Amanda and others},
  journal={Advances in neural information processing systems},
  volume={33},
  pages={1877--1901},
  year={2020}
}

@inproceedings{ye2025vera,
  title={Vera: Explainable video anomaly detection via verbalized learning of vision-language models},
  author={Ye, Muchao and Liu, Weiyang and He, Pan},
  booktitle={Proceedings of the Computer Vision and Pattern Recognition Conference},
  pages={8679--8688},
  year={2025}
}

@article{li2026vadtree,
  title={VADTree: Explainable training-free video anomaly detection via hierarchical granularity-aware tree},
  author={Li, Wenlong and Xu, Yifei and Rao, Yuan and Wang, Zhenhua and Deng, Shuiguang},
  journal={Advances in Neural Information Processing Systems},
  volume={38},
  pages={148372--148404},
  year={2026}
}

@article{cai2025training,
  title={Training-free group relative policy optimization},
  author={Cai, Yuzheng and Cai, Siqi and Shi, Yuchen and Xu, Zihan and Chen, Lichao and Qin, Yulei and Tan, Xiaoyu and Li, Gang and Li, Zongyi and Lin, Haojia and others},
  journal={arXiv preprint arXiv:2510.08191},
  year={2025}
}

@inproceedings{liu2024improved,
  title={Improved baselines with visual instruction tuning},
  author={Liu, Haotian and Li, Chunyuan and Li, Yuheng and Lee, Yong Jae},
  booktitle={Proceedings of the IEEE/CVF conference on computer vision and pattern recognition},
  pages={26296--26306},
  year={2024}
}

@article{agarwal2025gpt,
  title={gpt-oss-120b \& gpt-oss-20b model card},
  author={Agarwal, Sandhini and Ahmad, Lama and Ai, Jason and Altman, Sam and Applebaum, Andy and Arbus, Edwin and Arora, Rahul K and Bai, Yu and Baker, Bowen and Bao, Haiming and others},
  journal={arXiv preprint arXiv:2508.10925},
  year={2025}
}

@article{cai2024internlm2,
  title={Internlm2 technical report},
  author={Cai, Zheng and Cao, Maosong and Chen, Haojiong and Chen, Kai and Chen, Keyu and Chen, Xin and Chen, Xun and Chen, Zehui and Chen, Zhi and Chu, Pei and others},
  journal={arXiv preprint arXiv:2403.17297},
  year={2024}
}

@article{wang2025internvl3,
  title={Internvl3. 5: Advancing open-source multimodal models in versatility, reasoning, and efficiency},
  author={Wang, Weiyun and Gao, Zhangwei and Gu, Lixin and Pu, Hengjun and Cui, Long and Wei, Xingguang and Liu, Zhaoyang and Jing, Linglin and Ye, Shenglong and Shao, Jie and others},
  journal={arXiv preprint arXiv:2508.18265},
  year={2025}
}

@article{lin2026unified,
  title={A unified reasoning framework for holistic zero-shot video anomaly analysis},
  author={Lin, Dongheng and Qu, Mengxue and Han, Kunyang and Jiao, Jianbo and Jin, Xiaojie and Wei, Yunchao},
  journal={Advances in Neural Information Processing Systems},
  volume={38},
  pages={29713--29744},
  year={2026}
}

@inproceedings{shao2025eventvad,
  title={Eventvad: Training-free event-aware video anomaly detection},
  author={Shao, Yihua and He, Haojin and Li, Sijie and Chen, Siyu and Long, Xinwei and Zeng, Fanhu and Fan, Yuxuan and Zhang, Muyang and Yan, Ziyang and Ma, Ao and others},
  booktitle={Proceedings of the 33rd ACM International Conference on Multimedia},
  pages={2586--2595},
  year={2025}
}

@inproceedings{yang2024follow,
  title={Follow the rules: reasoning for video anomaly detection with large language models},
  author={Yang, Yuchen and Lee, Kwonjoon and Dariush, Behzad and Cao, Yinzhi and Lo, Shao-Yuan},
  booktitle={European Conference on Computer Vision},
  pages={304--322},
  year={2024},
  organization={Springer}
}

@inproceedings{radford2021learning,
  title={Learning transferable visual models from natural language supervision},
  author={Radford, Alec and Kim, Jong Wook and Hallacy, Chris and Ramesh, Aditya and Goh, Gabriel and Agarwal, Sandhini and Sastry, Girish and Askell, Amanda and Mishkin, Pamela and Clark, Jack and others},
  booktitle={International conference on machine learning},
  pages={8748--8763},
  year={2021},
  organization={PmLR}
}

@article{shi2023abnormal,
  title={Abnormal ratios guided multi-phase self-training for weakly-supervised video anomaly detection},
  author={Shi, Haoyue and Wang, Le and Zhou, Sanping and Hua, Gang and Tang, Wei},
  journal={IEEE Transactions on Multimedia},
  volume={26},
  pages={5575--5587},
  year={2023},
  publisher={IEEE}
}

@inproceedings{chen2024prompt,
  title={Prompt-enhanced multiple instance learning for weakly supervised video anomaly detection},
  author={Chen, Junxi and Li, Liang and Su, Li and Zha, Zheng-jun and Huang, Qingming},
  booktitle={Proceedings of the IEEE/CVF conference on computer vision and pattern recognition},
  pages={18319--18329},
  year={2024}
}

@inproceedings{huangex,
  title={Ex-VAD: Explainable Fine-grained Video Anomaly Detection Based on Visual-Language Models},
  author={Huang, Chao and Shi, Yushu and Wen, Jie and Wang, Wei and Xu, Yong and Cao, Xiaochun},
  booktitle={Forty-second International Conference on Machine Learning},
  year={2025}
}

@article{huang2025multimodal,
  title={Multimodal evidential learning for open-world weakly-supervised video anomaly detection},
  author={Huang, Chao and Huang, Weiliang and Jiang, Qiuping and Wang, Wei and Wen, Jie and Zhang, Bob},
  journal={IEEE Transactions on Multimedia},
  year={2025},
  publisher={IEEE}
}

@article{zhang2024holmes,
  title={Holmes-vad: Towards unbiased and explainable video anomaly detection via multi-modal llm},
  author={Zhang, Huaxin and Xu, Xiaohao and Wang, Xiang and Zuo, Jialong and Han, Chuchu and Huang, Xiaonan and Gao, Changxin and Wang, Yuehuan and Sang, Nong},
  journal={arXiv preprint arXiv:2406.12235},
  year={2024}
}

@article{yuksekgonul2025optimizing,
  title={Optimizing generative ai by backpropagating language model feedback},
  author={Yuksekgonul, Mert and Bianchi, Federico and Boen, Joseph and Liu, Sheng and Lu, Pan and Huang, Zhi and Guestrin, Carlos and Zou, James},
  journal={Nature},
  volume={639},
  number={8055},
  pages={609--616},
  year={2025},
  publisher={Nature Publishing Group UK London}
}

@article{xiao2024verbalized,
  title={Verbalized machine learning: Revisiting machine learning with language models},
  author={Xiao, Tim Z and Bamler, Robert and Sch{\"o}lkopf, Bernhard and Liu, Weiyang},
  journal={arXiv preprint arXiv:2406.04344},
  year={2024}
}

@article{meng2025audio,
  title={Audio-visual collaborative learning for weakly supervised video anomaly detection},
  author={Meng, Jingke and Tian, Huilin and Lin, Ge and Hu, Jian-Fang and Zheng, Wei-Shi},
  journal={IEEE Transactions on Multimedia},
  year={2025},
  publisher={IEEE}
}

@inproceedings{lv2023unbiased,
  title={Unbiased multiple instance learning for weakly supervised video anomaly detection},
  author={Lv, Hui and Yue, Zhongqi and Sun, Qianru and Luo, Bin and Cui, Zhen and Zhang, Hanwang},
  booktitle={Proceedings of the IEEE/CVF conference on computer vision and pattern recognition},
  pages={8022--8031},
  year={2023}
}

@article{tang2024hawk,
  title={Hawk: Learning to understand open-world video anomalies},
  author={Tang, Jiaqi and Lu, Hao and Wu, Ruizheng and Xu, Xiaogang and Ma, Ke and Fang, Cheng and Guo, Bin and Lu, Jiangbo and Chen, Qifeng and Chen, Ying-Cong},
  journal={Advances in Neural Information Processing Systems},
  volume={37},
  pages={139751--139785},
  year={2024}
}

@inproceedings{wu2024vadclip,
  title={Vadclip: Adapting vision-language models for weakly supervised video anomaly detection},
  author={Wu, Peng and Zhou, Xuerong and Pang, Guansong and Zhou, Lingru and Yan, Qingsen and Wang, Peng and Zhang, Yanning},
  booktitle={Proceedings of the AAAI conference on artificial intelligence},
  volume={38},
  number={6},
  pages={6074--6082},
  year={2024}
}

@inproceedings{li2022scale,
  title={Scale-aware spatio-temporal relation learning for video anomaly detection},
  author={Li, Guoqiu and Cai, Guanxiong and Zeng, Xingyu and Zhao, Rui},
  booktitle={European Conference on Computer Vision},
  pages={333--350},
  year={2022},
  organization={Springer}
}

@inproceedings{zhou2023dual,
  title={Dual memory units with uncertainty regulation for weakly supervised video anomaly detection},
  author={Zhou, Hang and Yu, Junqing and Yang, Wei},
  booktitle={Proceedings of the AAAI Conference on Artificial Intelligence},
  volume={37},
  number={3},
  pages={3769--3777},
  year={2023}
}

@inproceedings{tian2021weakly,
  title={Weakly-supervised video anomaly detection with robust temporal feature magnitude learning},
  author={Tian, Yu and Pang, Guansong and Chen, Yuanhong and Singh, Rajvinder and Verjans, Johan W and Carneiro, Gustavo},
  booktitle={Proceedings of the IEEE/CVF international conference on computer vision},
  pages={4975--4986},
  year={2021}
}

@inproceedings{rai2024video,
  title={Video Anomaly Detection via Spatio-Temporal Pseudo-Anomaly Generation: A Unified Approach},
  author={Rai, Ayush K and Krishna, Tarun and Hu, Feiyan and Drimbarean, Alexandru and McGuinness, Kevin and Smeaton, Alan F and O'connor, Noel E},
  booktitle={Proceedings of the IEEE/CVF Conference on Computer Vision and Pattern Recognition},
  pages={3887--3899},
  year={2024}
}

@inproceedings{ristea2024self,
  title={Self-distilled masked auto-encoders are efficient video anomaly detectors},
  author={Ristea, Nicolae-C and Croitoru, Florinel-Alin and Ionescu, Radu Tudor and Popescu, Marius and Khan, Fahad Shahbaz and Shah, Mubarak and others},
  booktitle={Proceedings of the IEEE/CVF conference on computer vision and pattern recognition},
  pages={15984--15995},
  year={2024}
}

@inproceedings{cai2026headhunt,
  title={HeadHunt-VAD: Hunting Robust Anomaly-Sensitive Heads in MLLM for Tuning-Free Video Anomaly Detection},
  author={Cai, Zhaolin and Li, Fan and Zheng, Ziwei and Bi, Haixia and He, Lijun},
  booktitle={Proceedings of the AAAI Conference on Artificial Intelligence},
  volume={40},
  number={24},
  pages={19835--19843},
  year={2026}
}

@article{li2025video,
  title={Video-Level Language-Driven Video-Based Visible-Infrared Person Re-Identification},
  author={Li, Shuang and Leng, Jiaxu and Kuang, Changjiang and Tan, Mingpi and Gao, Xinbo},
  journal={IEEE Transactions on Information Forensics and Security},
  year={2025},
  publisher={IEEE}
}

@article{li2023logical,
  title={Logical relation inference and multiview information interaction for domain adaptation person re-identification},
  author={Li, Shuang and Li, Fan and Li, Jinxing and Li, Huafeng and Zhang, Bob and Tao, Dapeng and Gao, Xinbo},
  journal={IEEE Transactions on Neural Networks and Learning Systems},
  volume={35},
  number={10},
  pages={14770--14782},
  year={2023},
  publisher={IEEE}
}

@article{li2025shape,
  title={Shape-centered representation learning for visible-infrared person re-identification},
  author={Li, Shuang and Leng, Jiaxu and Gan, Ji and Mo, Mengjingcheng and Gao, Xinbo},
  journal={Pattern Recognition},
  pages={111756},
  year={2025},
  publisher={Elsevier}
}
\bibliographystyle{sty/icml2026}

%%%%%%%%%%%%%%%%%%%%%%%%%%%%%%%%%%%%%%%%%%%%%%%%%%%%%%%%%%%%%%%%%%%%%%%%%%%%%%%
%%%%%%%%%%%%%%%%%%%%%%%%%%%%%%%%%%%%%%%%%%%%%%%%%%%%%%%%%%%%%%%%%%%%%%%%%%%%%%%
% APPENDIX
%%%%%%%%%%%%%%%%%%%%%%%%%%%%%%%%%%%%%%%%%%%%%%%%%%%%%%%%%%%%%%%%%%%%%%%%%%%%%%%
%%%%%%%%%%%%%%%%%%%%%%%%%%%%%%%%%%%%%%%%%%%%%%%%%%%%%%%%%%%%%%%%%%%%%%%%%%%%%%%
\newpage
\appendix
\onecolumn

\section{Theoretical Analysis}
\label{sec:app_theory}
Following~\cite{brown2020language}, we provide a first-order analysis showing that updating linguistic experience can serve as an approximate surrogate for parameter updates. LRPO keeps the VLM parameters fixed at $\theta_0$ and modulates the conditional output distribution $\pi_{\theta_0}(o\mid C)$ by editing the experience text, where $C=[P,\tilde{V},\mathcal{E}(\tilde{V})]$. Let $\mathbf{h}(C)$ denote the continuous representation of the context (e.g., input embeddings/hidden states). When the experience is updated from $\mathcal{E}$ to $\mathcal{E}'$, the context becomes $C'=[P,\tilde{V},\mathcal{E}'(\tilde{V})]$, inducing a perturbation $\Delta\mathbf{h}=\mathbf{h}(C')-\mathbf{h}(C)$. Under a local linearization assumption, for any output $o$ we have the first-order approximation
\begin{equation}
\log \pi_{\theta_0}(o\mid C')-\log \pi_{\theta_0}(o\mid C)\approx
\nabla_{\mathbf{h}}\log \pi_{\theta_0}(o\mid C)^{\top}\Delta\mathbf{h}.
\label{eq:ctx_taylor}
\end{equation}
Similarly, a small parameter update $\Delta\theta$ also induces a first-order change in $\log \pi_{\theta}(o\mid C)$. Therefore, in a first-order sense, the context perturbation $\Delta\mathbf{h}$ caused by experience editing can induce a distribution shift analogous to that of a small parameter update $\Delta\theta$, thereby changing the relative propensity of different outputs. This offers theoretical intuition for why LRPO can effectively steer the model distribution without updating any parameters.

We next interpret why LRPO can iteratively improve experience from a group-relative optimization perspective~\cite{shao2024deepseekmath}. For the same sample $(\tilde{V}^{(j)},Y^{(j)})$, we draw a group of outputs $\{o_k^{(j)}\}_{k=1}^{G}$ under the old experience $\mathcal{E}^{(j)}_{\mathrm{old}}$ and obtain rewards $\{r_k^{(j)}\}_{k=1}^{G}$, where $o_k^{(j)}\sim \pi_{\theta_0}(\cdot\mid \tilde{V}^{(j)},P,\mathcal{E}^{(j)}_{\mathrm{old}})$. The within-group normalized advantage is defined as
$\hat{A}_k^{(j)}=\frac{r_k^{(j)}-\mathrm{mean}(\{r_i^{(j)}\})}{\mathrm{std}(\{r_i^{(j)}\})}$.
For a candidate experience $\mathcal{E}^{(j)}$, we define the sequence-level likelihood ratio
$\rho_k^{(j)}=\frac{\pi_{\theta_0}(o_k^{(j)}\mid \tilde{V}^{(j)},P,\mathcal{E}^{(j)})}{\pi_{\theta_0}(o_k^{(j)}\mid \tilde{V}^{(j)},P,\mathcal{E}^{(j)}_{\mathrm{old}})}$,
and write the group-relative objective as
\begin{equation}
\begin{aligned}
J_{\mathrm{LRPO}}
=\mathbb{E}\Bigg[
\frac{1}{G}\sum_{k=1}^{G}
\min\Big(\rho_k^{(j)}\hat{A}_k^{(j)},\;
\mathrm{clip}(\rho_k^{(j)},1-\epsilon,1+\epsilon)\hat{A}_k^{(j)}\Big)
-\beta\,\mathrm{KL}\!\left(\pi_{\theta_0}(\cdot\mid \tilde{V}^{(j)},P,\mathcal{E}^{(j)})\;\|\;\pi_{\theta_0}(\cdot\mid \tilde{V}^{(j)},P)\right)
\Bigg].
\end{aligned}
\label{eq:grp_improve}
\end{equation}
Here the clipping term and the KL regularizer jointly bound the per-step distribution change, keeping each experience update in a ``small-step'' regime and thus enabling stable policy improvement under the strong prior induced by the frozen base model $\pi_{\theta_0}$.

\section{Dataset Statistics}
\label{sec:app_dataset_stats}
\revise{Because test videos in VAD benchmarks usually contain highly imbalanced normal and abnormal frames, quantifying this distribution is important for clarifying the experimental setting and improving evaluation transparency. Therefore, we summarize the normal/abnormal video counts and, when frame-level annotations are available, the corresponding frame distributions in Table~\ref{tab:dataset_frame_stats}. For UCF-Crime and XD-Violence, the training splits are annotated only at the video level, whereas frame-level annotations are provided for testing; therefore, their training statistics are reported only by video count. In contrast, UBnormal provides frame-level annotations for all splits, allowing frame statistics to be reported for the train, validation, and test sets.}
\begin{table}[h]
\centering
\caption{\protect\revise{Dataset-level video counts and frame-level normal/abnormal statistics.}}
\label{tab:dataset_frame_stats}
\begingroup\color{revisioncolor}
\resizebox{\linewidth}{!}{
\begin{tabular}{l l r r r r r r}
\toprule
\textbf{Dataset} & \textbf{Split} & \textbf{\# Normal Videos} & \textbf{\# Abnormal Videos} & \textbf{\# Total Videos} & \textbf{\# Normal Frames} & \textbf{\# Abnormal Frames} & \textbf{Abnormal Frame Ratio} \\
\midrule
\multirow{2}{*}{UCF-Crime} & Train & 800 & 810 & 1,610 & N/A & N/A & N/A \\
& Test & 150 & 140 & 290 & 1,027,477 & 84,331 & 7.59\% \\
\midrule
\multirow{2}{*}{XD-Violence} & Train & 2,049 & 1,905 & 3,954 & N/A & N/A & N/A \\
& Test & 300 & 500 & 800 & 1,806,004 & 529,797 & 22.68\% \\
\midrule
\multirow{3}{*}{UBnormal} & Train & 186 & 82 & 268 & 90,860 & 25,227 & 21.73\% \\
& Validation & 26 & 38 & 64 & 14,237 & 13,938 & 49.47\% \\
& Test & 53 & 158 & 211 & 42,790 & 49,850 & 53.81\% \\
\bottomrule
\end{tabular}
}
\endgroup
\end{table}

% \section{Additional experimental results}
% \paragraph{Why naive instruction tuning fails under weak supervision.}
% Tab.~\ref{tab:sft} reports Qwen3-VL-2B-Instruct under the same weak video-level supervision used in our setting. A short SFT (1 epoch) yields a large AP gain over the base model, but performance consistently drops as training continues (2-3 epochs), indicating rapid overfitting to dataset-specific shortcuts and noisy weak labels. This trend suggests that directly optimizing model weights with weak supervision can be unstable and brittle: it improves in-domain AP early on, yet progressively harms generalization as the model memorizes spurious correlations rather than learning robust anomaly reasoning. In contrast, our LRPO keeps the VLM frozen and iteratively refines a linguistic experience library with feedback, which mitigates such weight-level overfitting while retaining the foundation model prior.
% \label{sec:app_experiment}
% \input{tables/tab_sft}

\section{Additional Implementation Details}
\label{sec:app_impl_details}
% For the anomaly preference reward, we use InternVL3\_5-38B to generate perturbed preference texts as hard negatives, with $H=3$ per sample.
% To ensure stable and lightweight experience editing, we restrict the optimizer to perform at most $3$ experience operations per update step.
% For the anomaly temporal dependency reward, we set the suppression coefficient to $\mu=0.8$.
% During inference, we follow~\cite{ye2025vera} to incorporate both scene context and temporal context into the reasoning process.
For the anomaly preference reward, we use InternVL3\_5-38B to generate perturbed preference texts as hard negatives, with $H=3$ per sample, and set the temperature to $\tau=0.1$.
To ensure stable and lightweight experience editing, we restrict the optimizer to perform at most $3$ experience operations per update step.
\revise{We cap the general experience repository at $|\mathcal{E}^{\mathrm{gen}}|\le 30$ and require each retained item to describe transferable anomaly judgment principles rather than sample specific details, which keeps the repository compact and helps reduce redundancy and uncontrolled drift.}
For the anomaly temporal dependency reward, we set the suppression coefficient to $\mu=0.8$ and use $\epsilon_0=0.05$ and $\epsilon_1=0.20$ for the soft gate in Eq.~(\ref{eq:r_tem_group}).
During inference, we follow~\cite{ye2025vera} to incorporate both scene context and temporal context into the reasoning process.
\revise{Overall, although LRPO uses a three-stage pipeline, the workflow remains lightweight in practice, with the complete training process taking only about $2.3$ hours.}

\section{Experience Library Showcases}
\label{sec:app_expe}
\subsection{Showcase of General Experience Learned by LRPO}
We provide the complete list of general anomaly preference experiences learned by LRPO. These experiences are learned on the XD-Violence dataset using only $2.5\%$ of the training set. Each item is a natural-language rule that captures transferable decision criteria across scenarios, mainly covering temporal constraints, evidence composition, and conflict-resolution priorities, and can be directly injected as a contextual prior at inference time.
% 我们在此给出 LRPO 学习到的通用异常偏好经验条目（完整列表）。这些经验是在 XD-Violence 数据集上仅使用 $2.5\%$ 训练数据学习得到的：每条条目以自然语言规则形式刻画跨场景可迁移的判别准则，主要覆盖时间约束、证据组合与冲突消解优先级，并可在推理时作为上下文先验直接注入。

\small
\begin{enumerate}
\item Active fighting requires aggressive physical contact persisting $\ge$3 consecutive frames; weapons do not preclude Fighting if intent to harm is clear, otherwise apply Abuse when coercive handling appears without contact.
\item When weapons appear, also label Shooting if a bright or muzzle flash is followed within 3 frames by injury, distress, or protective reaction, even when the weapon is not yet visible.
\item When someone is lying on the ground, label Abuse if coercive or restraint cues appear within 2-3 frames and no weapon is visible; otherwise defer to Shooting.
\item When law-enforcement appears before civilians, label Riot if within 5\,s at least two aggression cues emerge, or if a single cue persists $\ge$2\,s after the police appearance.
\item Emotional distress alone is insufficient; if distress co-occurs with discharge or protective reaction within 2-3 frames, label Shooting. If distress appears with a weapon aimed at a vulnerable person without discharge, label Abuse.
\item When multiple cues appear, label each only if its rule holds in separate segments; resolve overlaps with priority: Explosion $>$ Shooting $>$ Riot $>$ Fighting $>$ Abuse $>$ Car-accident $>$ Normal.
\item When restrained and another actor forcibly applies an object or substance (e.g., pours, sprays, pushes) onto the person, and visible distress follows within $\le$3 frames, label Abuse.
\item When reduced visibility persists, scan early, middle, and late checkpoints; if no high-severity or restraint/forced-handling cues appear and the setting looks benign, label Normal.
\item When only aftermath artifacts like lingering smoke or debris appear, no high-severity explosion cue occurred in the prior 5\,seconds, and artifacts show no rapid growth, label Normal; otherwise treat as ongoing Explosion.
\item When an on-scene authority figure visibly separates combatants, treat as supporting evidence for genuine Fighting only if aggressive contact cues already exist; ignore in staged scenes without hostile intent.
\item When a crowd shifts from passive to aggressive, label Riot if two of: density rise, chanting, aggression cues (projectiles, tear-gas, protest symbols, fire, smoke, flares) appear within 5 seconds; cue persisting $\ge$1 frame.
\item When a vulnerable individual is restrained by a dominant aggressor and visible distress appears, label Abuse; weapon presence does not downgrade unless a discharge flash or weapon-caused injury occurs within two frames.
\item When a sudden bright flash is large and sustained, and within the next few frames smoke, fire, debris, shockwave, or damage appear, prioritize Explosion over Shooting, even if a weapon is present.
\item When reduced visibility persists, scan the next few seconds for abrupt motion, sudden light or sound cues, emergency vehicles, or rapid crowd shifts; if detected, invoke the matching specific anomaly rule.
\item When weapons appear but no discharge cue in 2-3 frames, run the flash-explosion scan; if Explosion is detected, label Explosion, else do not label Shooting and monitor or consider Abuse.
\item When a sudden bright flash is detected, scan the next five frames for explosion cues (smoke, debris, fire, shockwave, damage); if found, label Explosion, overriding other cues.
\item When law-enforcement and civilians confront and officers wear riot gear, treat megaphones, banners, chanting, raised fists, coordinated gestures, or density surge as Riot cues persisting $\ge$1 frame.
\item When a vulnerable individual shows visible injury and a dominant aggressor is present, require a coercive action (e.g., grip, push) within $\le$2 frames followed by distress within $\le$3 frames to label Abuse.
\item When a sudden bright flash occurs without explosion cues, label Shooting only if a weapon-related cue appears or injury/distress directly follows the flash within 2-3 frames; otherwise continue scanning.
\item When a sudden bright flash (e.g., flare or fireworks) occurs with law-enforcement and a hostile crowd, and any aggression cue appears within 3\,seconds, label Riot with elevated confidence.
\item When an anomaly cue appears and any label is assigned, always continue scanning remaining frames for independent cues of other categories, applying each rule and resolving overlaps by priority.
\item When a vehicle abruptly changes motion, and within five seconds any of: damage, smoke, fire, debris, emergency-response lights, or clear traffic disruption (stopped vehicles, congestion, lane blockage) appear, label Car-accident, overriding lower-severity cues.
\item When weapons appear, prioritize Shooting over Fighting unless the weapon is sport equipment in a recognized sport context; then evaluate aggressive contact under the Fighting rule if no discharge or injury cues.
\item When weapons appear without discharge, injury, or protective reaction in 3-5 frames, prioritize Abuse labeling if restraint or distress cues exist; only label Shooting with a discharge flash or direct weapon-injury link.
\item When an anomaly cue appears and no high-severity cues are detected across early, middle, and late segments, and behavior is ordinary for the setting, assign the Normal label.
\item When a sudden bright flash is large and smoke plumes, flames, or fireballs persist for three or more frames, especially with shockwave or debris, label Explosion regardless of weapon.
\item When law-enforcement appears before civilians wearing riot gear, label Riot if two aggression cues span $\ge$3\,seconds, boosting confidence as strong temporal evidence.
\end{enumerate}
\normalsize

\subsection{Showcase of Selected Scenario Experience}
% 我们在此节选并展示部分场景经验条目。每条经验以自然语言规则刻画特定场景下的异常线索、触发条件与判别边界，可在推理时作为可检索的上下文先验注入。
We provide an excerpt of the scenario experience entries. Each entry is a natural-language rule that specifies context-dependent anomaly cues, triggering conditions, and decision boundaries, and can be retrieved and injected as an in-context prior during inference.
\begin{itemize}
  \item \textbf{Explosion.}
  \begin{itemize}
    \item When in an urban combat scene, if a person aims and fires a weapon from a rooftop, causing an explosion inside a building, you should judge it as a shooting and explosion anomaly.
    \item When in a combat scene, if gunfire and explosions are visible, you should judge it as a shooting and explosion anomaly.
    \item When in a battlefield scene, if soldiers are seen aiming and firing weapons at surrendering individuals, and there are signs of explosions and fire, you should judge it as a shooting and explosion anomaly.
    \item When in an industrial or high-tension scene, if an explosion occurs (fire, smoke, debris), you should judge it as an explosion anomaly.
    % \item When in an outdoor scene, if a large explosion with fire and smoke is visible, you should judge it as an explosion anomaly.
    % \item When in a street scene, if a vehicle or object explodes producing fire and thick smoke, you should judge it as an explosion anomaly.
    % \item When in an urban street scene, if an explosive blast is followed by panic, running crowd, or debris, you should judge it as an explosion anomaly.
    % \item When in a building interior, if a sudden bright flash is followed by flames or smoke, you should judge it as an explosion anomaly.
    % \item When in a night scene, if an explosion illuminates the area and is followed by fire, you should judge it as an explosion anomaly.
    % \item When in a crowded area, if an explosion occurs and people scatter or fall, you should judge it as an explosion anomaly.
  \end{itemize}

  \item \textbf{Shooting.}
  \begin{itemize}
    \item When in a street scene, if a person points a gun at another person and fires, you should judge it as a shooting anomaly.
    \item When in an indoor scene, if a person is holding a gun and a muzzle flash is visible, you should judge it as a shooting anomaly.
    \item When in a combat scene, if people are aiming rifles and firing, you should judge it as a shooting anomaly.
    \item When in a robbery-like scene, if a weapon is brandished and a shot is fired, you should judge it as a shooting anomaly.
    % \item When in a confrontation scene, if a person fires a handgun and the target shows distress or falls, you should judge it as a shooting anomaly.
    % \item When in a public space, if a gun is fired and bystanders react by ducking or running, you should judge it as a shooting anomaly.
    % \item When in a police-related scene, if officers or civilians discharge firearms, you should judge it as a shooting anomaly.
    % \item When in an outdoor scene, if gunfire occurs and people take cover, you should judge it as a shooting anomaly.
    % \item When in a close-range altercation, if a firearm is used and a shot is fired, you should judge it as a shooting anomaly.
    % \item When in a conflict scene, if automatic gunfire is present with visible muzzle flash, you should judge it as a shooting anomaly.
    % \item When in a street confrontation, if a person aims a weapon at another and the target recoils, you should judge it as a shooting anomaly.
    % \item When in an indoor corridor, if a gun is fired and smoke or flash is visible, you should judge it as a shooting anomaly.
    % \item When in a vehicle scene, if a person shoots from a car and others react, you should judge it as a shooting anomaly.
    % \item When in a battlefield scene, if multiple shooters fire toward a crowd or group, you should judge it as a shooting anomaly.
    % \item When in a protest or riot-like scene, if a gun is fired (not fireworks) and panic follows, you should judge it as a shooting anomaly.
  \end{itemize}

  \item \textbf{Riot.}
  \begin{itemize}
    \item When in a street protest scene, if a crowd confronts law enforcement and aggressive actions occur, you should judge it as a riot anomaly.
    \item When in a public disturbance scene, if people are throwing objects or clashing with police, you should judge it as a riot anomaly.
    \item When in an outdoor crowd scene, if a mob becomes violent and pushes or attacks, you should judge it as a riot anomaly.
    \item When in a protest scene, if tear gas, shields, or riot gear appears with hostile crowd behavior, you should judge it as a riot anomaly.
    % \item When in a large gathering, if people surge, chant, and engage in aggressive confrontation, you should judge it as a riot anomaly.
    % \item When in a city street, if multiple individuals attack others amid a crowd, you should judge it as a riot anomaly.
    % \item When in a riot scene, if fires, smoke, or debris appear with crowd aggression, you should judge it as a riot anomaly.
    % \item When in a law-enforcement scene, if officers and civilians clash and the crowd becomes chaotic, you should judge it as a riot anomaly.
    % \item When in a protest scene, if barricades, projectiles, and police confrontation co-occur, you should judge it as a riot anomaly.
    % \item When in an outdoor plaza, if the crowd turns violent and people are harmed, you should judge it as a riot anomaly.
    % \item When in a street scene, if a crowd attacks vehicles or property while confronting authorities, you should judge it as a riot anomaly.
    % \item When in a nighttime crowd scene, if flares or fires appear with aggressive mob behavior, you should judge it as a riot anomaly.
    % \item When in a public unrest scene, if fighting spreads across multiple people in a crowd, you should judge it as a riot anomaly.
    % \item When in a confrontation scene, if riot police appear and civilians throw objects or rush forward, you should judge it as a riot anomaly.
    % \item When in a protest environment, if coordinated aggressive gestures and density surge appear, you should judge it as a riot anomaly.
  \end{itemize}

  \item \textbf{Fighting.}
  \begin{itemize}
    \item When in a street scene, if two or more people are punching or kicking each other, you should judge it as a fighting anomaly.
    \item When in an indoor scene, if a physical fight breaks out with sustained aggressive contact, you should judge it as a fighting anomaly.
    \item When in a public place, if multiple individuals engage in a brawl, you should judge it as a fighting anomaly.
    \item When in a hallway scene, if people grapple and strike each other, you should judge it as a fighting anomaly.
    % \item When in a crowded scene, if a scuffle escalates into sustained punching or kicking, you should judge it as a fighting anomaly.
    % \item When in a shop or market, if people physically assault each other, you should judge it as a fighting anomaly.
    % \item When in a domestic scene, if a dispute turns into visible physical combat, you should judge it as a fighting anomaly.
    % \item When in a street corner, if an argument leads to repeated blows, you should judge it as a fighting anomaly.
    % \item When in a sports-like setting, if the context is not a recognized sport and aggressive contact occurs, you should judge it as a fighting anomaly.
    % \item When in a parking area, if two people push, punch, or wrestle on the ground, you should judge it as a fighting anomaly.
    % \item When in a transit station, if a crowd forms around a physical altercation, you should judge it as a fighting anomaly.
    % \item When in an outdoor scene, if one person repeatedly strikes another and others intervene, you should judge it as a fighting anomaly.
  \end{itemize}

  \item \textbf{Abuse.}
  \begin{itemize}
    \item When in a domestic scene, if a person forcibly restrains or drags another person and distress is visible, you should judge it as an abuse anomaly.
    \item When in an indoor scene, if a vulnerable individual is restrained and coerced, you should judge it as an abuse anomaly.
    \item When in a street scene, if an aggressor holds down a person and applies coercive force, you should judge it as an abuse anomaly.
    \item When in a room scene, if a person is being tied up or held captive, you should judge it as an abuse anomaly.
    % \item When in a confined space, if coercive handling occurs without clear weapon discharge, you should judge it as an abuse anomaly.
    % \item When in a scene with a dominant aggressor and a vulnerable victim, if forceful control and distress co-occur, you should judge it as an abuse anomaly.
    % \item When in a kidnapping-like scene, if a person is forced into a vehicle or dragged away, you should judge it as an abuse anomaly.
    % \item When in an indoor setting, if someone is pinned or restrained while being threatened, you should judge it as an abuse anomaly.
  \end{itemize}

  \item \textbf{Car accident.}
  \begin{itemize}
    \item When in a road scene, if vehicles collide and damage is visible, you should judge it as a car accident anomaly.
    \item When in a highway scene, if a vehicle crashes and smoke or debris appears, you should judge it as a car accident anomaly.
    \item When in an intersection, if a sudden collision causes traffic disruption, you should judge it as a car accident anomaly.
    \item When in a street scene, if a car hits an object or another vehicle and people react, you should judge it as a car accident anomaly.
    % \item When in a road scene, if a vehicle flips or rolls over, you should judge it as a car accident anomaly.
    % \item When in a driving scene, if a crash is followed by fire or smoke, you should judge it as a car accident anomaly.
    % \item When in an outdoor traffic scene, if a pileup occurs with visible wreckage, you should judge it as a car accident anomaly.
    % \item When in a roadway, if a vehicle abruptly stops with damage and emergency response appears, you should judge it as a car accident anomaly.
    % \item When in a parking or roadside scene, if a vehicle impact leads to visible damage and distress, you should judge it as a car accident anomaly.
    % \item When in a traffic scene, if a collision results in debris scattered on the road, you should judge it as a car accident anomaly.
  \end{itemize}

  \item \textbf{Normal.}
  \begin{itemize}
    \item When in a sports competition scene, if the actions are consistent with play and no harm cues appear, you should judge it as normal.
    \item When in a stage performance scene, if violence-like gestures appear but are clearly choreographed, you should judge it as normal.
    \item When in a training or drill scene, if weapons appear without hostile intent or discharge cues, you should judge it as normal.
    \item When in a construction scene, if sparks or bright lights occur as part of work activity, you should judge it as normal.
    % \item When in an outdoor celebration scene, if flashes are consistent with fireworks and no distress follows, you should judge it as normal.
    % \item When in a crowded street scene, if people are moving normally without aggression cues, you should judge it as normal.
    % \item When in a workplace scene, if loud activity or fast motion is part of routine operations, you should judge it as normal.
    % \item When in a public transit scene, if crowd density is high but behavior remains orderly, you should judge it as normal.
    % \item When in a smoke-like scene, if smoke is static and no fire, damage, or panic is present, you should judge it as normal.
    % \item When in an indoor gathering, if raised voices occur without physical aggression, you should judge it as normal.
  \end{itemize}
\end{itemize}

\section{Additional qualitative analyses}
\label{sec:app_qual}

\begin{figure}[!htb]
\centerline{\includegraphics[width=0.6\linewidth, keepaspectratio]{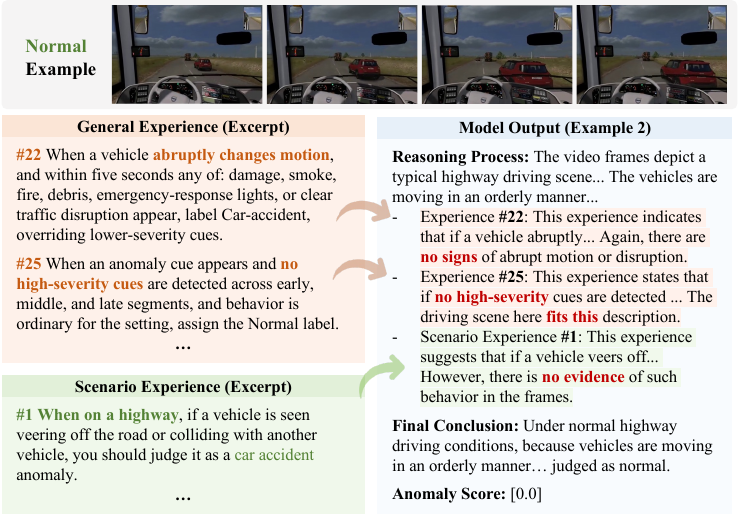}}
% \centerline{\includegraphics[keepaspectratio]{assets/vis_app_qual.pdf}}
% \caption{Qualitative analysis.}
% \caption{Qualitative example of LRPO, showing how general experience and scenario experience jointly support interpretable anomaly reasoning.}
% \caption{Qualitative example of LRPO, showing how general and scenario experiences jointly align anomaly reasoning with risk preferences.}
% \caption{Qualitative example on video ``v=251\_\_\_mEwZA'' from XD-Violence, showing how general and scenario experiences jointly align anomaly reasoning with risk preferences.}
% \caption{Qualitative example on video ``v=251\_\_\_mEwZA'', showing how LRPO leverages general and scenario experiences to produce anomaly-preference-aligned reasoning.}
\caption{Qualitative example on video ``v=y7JEq-kf2I''. General experience provides transferable criteria (e.g., checking abrupt motion and high-severity cues), while scenario experience supplies scene-specific triggers and boundary calibration for the highway-driving context, leading to a consistent normal conclusion.}
% \caption{Qualitative example on video ``v=251\_\_\_mEwZA'' where injecting learned general experience and retrieved scenario experience leads to reasoning aligned with anomaly preferences.}
\label{fig:vis_app_qual}
% \vspace{-8pt}
\end{figure}

Figure~\ref{fig:vis_app_qual} provides a normal-case example illustrating how LRPO combines learned general experience with retrieved scenario experience to support preference-aligned reasoning.
The general experience items (e.g., \#22 and \#25) act as transferable decision criteria, prompting the model to verify whether abrupt motion changes, high-severity cues, or other accident evidence emerge across temporally separated segments.
Meanwhile, the retrieved scenario experience (e.g., \#1) supplies a context-dependent trigger rule for highway driving, calibrating what constitutes a car-accident boundary under this scene.
Since the frames exhibit orderly driving and lack the required evidence (no abrupt motion, disruption, or damage cues), the model consistently rules out anomalies and outputs a normal conclusion with a low anomaly score.
Overall, this case suggests that general experience provides reusable anomaly preferences and conflict-resolution principles, while scenario experience anchors these preferences to scene-specific boundaries, jointly improving reasoning consistency.

\begin{figure*}[ht]
% \vspace{-2pt}

\centering
\includegraphics[width=0.75\textwidth]{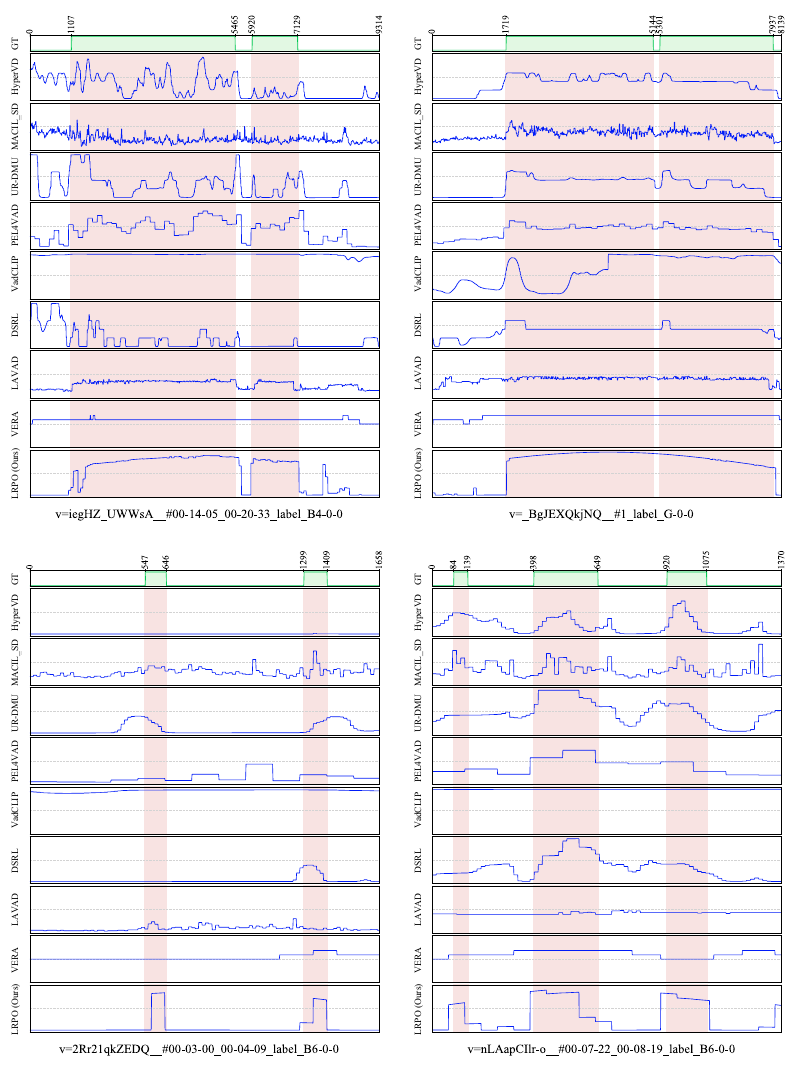} % Reduce the figure size so that it is slightly narrower than the column.
% \vspace{-20pt}
\caption{
    Additional visualization of frame-level anomaly detection results on XD-Violence. The gray horizontal line at 0.5 indicates the anomaly detection threshold.
    }
\label{fig:vis_app_score}
\end{figure*}

Figure~\ref{fig:vis_app_score} visualizes several frame-level anomaly score curves on the XD-Violence dataset.
We compare LRPO with representative tuning-based methods, including HyperVD~\cite{zhou2024learning}, MACIL\_SD~\cite{yu2022modality}, UR-DMU~\cite{zhou2023dual}, PEL4VAD~\cite{pu2024learning}, and DSRL~\cite{leng2024beyond}, as well as tuning-free methods such as LAVAD~\cite{zanella2024harnessing} and VERA~\cite{ye2025vera}.
Across these examples, LRPO produces score trajectories that better align with the temporal extent of ground-truth anomalous intervals: it yields clearer separation between normal and abnormal regions, assigns higher confidence within anomalous segments while suppressing spurious spikes in normal frames, and exhibits more stable temporal consistency near boundaries.
These qualitative results suggest that injecting and iteratively optimizing anomaly experience helps LRPO form more reliable temporally grounded anomaly evidence, complementing both training-based detectors and existing tuning-free approaches.

\end{document}